\documentclass[opre,nonblindrev]{informs3}

\DoubleSpacedXI 

\usepackage{enumitem}

\newenvironment{roster}
 {\begin{enumerate}[label=(\roman*)]}
 {\end{enumerate}}

\usepackage{hyperref}
\hypersetup{
    colorlinks,
    citecolor=black,
    filecolor=black,
    linkcolor=black,
    urlcolor=black
}

\usepackage{tikz}
\usetikzlibrary{fit,tikzmark}

\newcommand\drawCodeBox[2]{%
  \begin{tikzpicture}[remember picture,overlay]
    \coordinate (start) at ([yshift=1.7ex]pic cs:#1);
    \coordinate (end) at ([yshift=-0.6ex]pic cs:#2);
    \node[inner sep=2pt,draw=red,fit=(start) (end)] {};
  \end{tikzpicture}%
}

\usepackage[normalem]{ulem}

\usepackage{natbib}
 \bibpunct[, ]{(}{)}{,}{a}{}{,}%
 \def\bibfont{\small}%
\OneAndAHalfSpacedXI
\TheoremsNumberedThrough     
\ECRepeatTheorems
\EquationsNumberedThrough    

\begin{document}


\RUNAUTHOR{Zhao, Xie, and Luo}

\RUNTITLE{Sensitivity Analysis on pKG Hybrid Models with SV Estimation}

\TITLE{Sensitivity Analysis on Policy-Augmented Graphical Hybrid Models with Shapley Value Estimation}

\ARTICLEAUTHORS{%
\AUTHOR{Junkai Zhao}
\AFF{Antai College of Economics and  Management, Shanghai Jiao Tong University, Shanghai 200030, \EMAIL{zhaojunkai@sjtu.edu.cn}} 
\AUTHOR{Wei Xie\thanks{Corresponding authors.}}
\AFF{Department of Mechanical and Industrial Engineering, Northeastern University, Boston, 02115, \EMAIL{w.xie@northeastern.edu}}
\AUTHOR{Jun Luo\footnotemark[1]}
\AFF{Antai College of Economics and  Management, Shanghai Jiao Tong University, Shanghai 200030, \EMAIL{jluo\_ms@sjtu.edu.cn}}
\AFF{Data-Driven Management Decision Making Lab, Shanghai Jiao Tong University}
} 

\ABSTRACT{Driven by the critical challenges in biomanufacturing, including high complexity and high uncertainty, we propose a comprehensive and computationally efficient sensitivity analysis framework for general nonlinear policy-augmented knowledge graphical  (pKG) hybrid models that characterize the risk- and science-based understandings of underlying stochastic decision process mechanisms. The criticality of each input (i.e., random factors, policy parameters, and model parameters) is measured by applying Shapley value (SV) sensitivity analysis to pKG (called SV-pKG), accounting for process causal interdependences. 
To quickly assess the SV for heavily instrumented bioprocesses, we approximate their dynamics with linear Gaussian pKG models and improve the SV estimation efficiency by utilizing the linear Gaussian properties. In addition, we propose an effective permutation sampling method with TFWW transformation and variance reduction techniques, namely the quasi-Monte Carlo and antithetic sampling methods, to further improve the sampling efficiency and estimation accuracy of SV for both general nonlinear and linear Gaussian pKG models. 
Our proposed framework can benefit efficient interpretation and support stable optimal process control in biomanufacturing.
}




\maketitle

\section{Introduction}


Biomanufacturing plays an important role in supporting public health and economic growth. By 2015, more than 345 million patients worldwide had benefited from biopharmaceuticals for treating cancer, diabetes, and
many other diseases \citep{koca2023increasing}. Especially during the past few years, it has played a prominent role in responding to the threat of the COVID-19 pandemic. 
More than 4.8 billion people have been vaccinated, and data from the US Centers for Disease Control and Prevention (CDC) show that full vaccination decreases the risk of death by 12-fold for those over 50 years of age \citep{walsh2022biopharmaceutical}. The global biopharmaceutical market was valued at \$343 billion in 2021  and it is projected to increase to a revenue of \$478 billion by 2026 \citep{kamaludin2023biomanufacturing}. 

Despite the benefits and growing demand, existing biomanufacturing processes face critical challenges,
including high complexity, high uncertainty, and very limited data \citep{tellechea2022fast}. Biotherapeutics are manufactured in living organisms (e.g., cells) whose biological processes are
very complex and have high variability.
Also, the manufacturing process typically consists of multiple integrated unit operations, such as fermentation, centrifugation, and filtration, which makes it even more complex \citep{clomburg2017industrial}. The application of modern sensor technology enables the decision-makers to capture more states and 
dynamics information of the biomanufacturing process \citep{reardon2021practical}, which further scales up the dimension in the characterization and control of the biomanufacturing process.
Besides, since the biological life cycle is short and the production lead time is long, each of the biomanufacturing experiments is expensive and therefore, the amount of historical batch data is limited. Having less than 20 process observations often happens in biomanufacturing \citep{o2021hybrid}. 


Various process analytical technologies (PATs) and control strategies have been proposed to guide decision-making and control in the biomanufacturing
process. First-principle models employ well-developed process understanding in the form of biochemical equations, reaction mechanisms and constitutive relations \citep{hong2018challenges1, luo2021bioprocess, ye2014optimization}. 
However, such first-principle models are often unavailable for the integrated biomanufacturing process due to the inherent complexity of biological molecules and processes. Also, they typically overlook bioprocess intrinsic uncertainty, such as uncertainties from inputs (e.g., raw materials, actions and other uncontrolled
variables),
and model misspecification.
To overcome these issues, some data-driven PATs have been developed, e.g., using neural networks \citep{natarajan2021online} and Gaussian processes \citep{zeng2018constrained} as predictive models, in the field of biomanufacturing. 
Such data-driven methods enable decision-makers to build black-box metamodels based on real-world data \citep{ding2022sample}. However, they are less interpretable and often ignore prior knowledge of the mechanism information, which may lead to poor performance under conditions with limited data. 
Driven by these challenges and limitations, \cite{xie2022interpretable} introduce a Bayesian network hybrid model and sensitivity analysis, 
which can leverage existing mechanisms and learn from real-world process data. Meanwhile, \cite{zheng2022policy} propose an interpretable policy-augmented knowledge graphical (pKG) hybrid model to guide integrated biomanufacturing process control, which augments the Bayesian network hybrid model with control policy. These works build on the prior structural information from existing kinetic models and facilitate learning from real-world data, which overcomes the limitations of pure first-principle models and data-driven models \citep{zhao2023policy}.


The utility of a hybrid bioprocess model can be  further used to investigate which input variables contribute more signiﬁcantly to the outputs, enabling more effective efforts on these important attributes to control manufacturing performance. In order to quantify the relative importance of input variables or factors to the output, sensitivity analysis, 
which has been widely used for the analysis of parameter uncertainty in mathematical models and simulations \citep{lam2016robust, reich2023sensitivity,Xie_EJOR_2023}, 
can benefit the understanding of the stochastic model in biomanufacturing.

Existing sensitivity analysis approaches mostly focus on the black-box simulation model \citep{borgonovo2023sensitivity} or metamodel such as the Gaussian process regression model
\citep{morris2018decomposing}. One commonly used method is the variance-based method, 
which assesses the importance of
model inputs based on the expected reduction in the output performance estimation uncertainty
\citep{owen2013variance, borgonovo2016sensitivity}. Regression-based \citep{de2016estimation} and information-based \citep{wiesel2022measuring} methods are also developed
in the sensitivity analysis.
\cite{owen2014sobol} introduces a Shapley value (SV) sensitivity analysis method based on the concept in game theory. As SV-based analysis satisfies several appealing properties for interpretation and system dynamics analysis, it is now popular in many fields, including operations research \citep{singal2022shapley} 
and interpretable artificial intelligence 
\citep{lundberg2017unified1}. In this paper, we introduce a comprehensive SV-based sensitivity analysis approach for the pKG model in biomanufacturing to provide the risk- and science-based
understanding of the complex biomaufacturing process.

Several works have been proposed to conduct SV-based sensitivity analysis on the biomanufacturing process. In particular, \mbox{\cite{xie2022interpretable}} and \mbox{\cite{zheng2022policy}}
developed SV-based sensitivity analysis on linear Gaussian models to quantify the contributions of intermediate state 
 and action variables to the output of interest and derive the analytical formula of SV. 
For general nonlinear transition models of bioprocesses,  the SV analysis is not straightforward and may lack explicit formulas. In addition,  the impact of intermediate states and actions may not be the fundamental underlying variables that affect the operations of the biomanufacturing system.  Therefore, in this paper, we consider a more general structure of the pKG model and develop simulation-based framework to efficiently estimate SV in both general nonlinear and linear pKG models. We study the sensitivity analysis over three types of input variables, i.e., random factors, policy parameters and model parameters, which are the key driving factors of the final outputs. To be more  precise, 
random factors represent the key sources of variation in biomanufacturing, and their sensitivity analysis helps identify the most sensitive and high-risk operational components. Policy parameters define the control policy during production, with their SVs revealing key driving forces of process outcomes and associated risks. Model parameters, often estimated from limited real-world data, 
are important to capture the model uncertainty in the bioprocess, and their SVs can support more informative data collection during lab experiments.

As we consider so many input variables, one drawback of the exact calculation of SV is that its computation cost grows exponentially as the number of input variables increases. In fact, the number of input variables could be large in a bioprocess, especially when there are multiple units of operations.
For heavily instrumented biomanufacturing
processes, where online sensor monitoring technologies
are used to facilitate real-time process control, there may exist tens to thousands of time steps \citep{papathanasiou2019assisting, martagan2023merck}.
The resulting high computational cost of SV limits its use in integrated-process applications, including both laboratory experiments and online process control in biomanufacturing. 
First, if sensitivity analysis is time-consuming, additional holding time and setup cost may occur, which can be extremely high in bioprocess experiments \citep{martagan2023merck}. 
Second, inefficient estimation of sensitivity indices can delay the response to the source of variations and the bottleneck of the biomanufacturing system, which leads to adverse effects on quality consistency and productivity of the biopharmaceuticals~\citep{drobnjakovic2023current}.

In this paper,
we first consider the linear Gaussian pKG to approximate the behavior of the bioprocess, which can also improve the SV estimation efficiency in such heavily instrumented systems. We derive the analytical formula for the SV of random factors. As the SVs of policy parameters and model parameters still lack exact formulas even in the linear Gaussian pKG model, we design an efficient estimation algorithm for the SV sensitivity analysis over inputs, including policy parameters and model parameters, utilizing the structural properties of linear Gaussian pKG. With the calculation reusing, we show that our algorithm can reduce the computational cost by $O(H)$ for the predictive analysis and $O(H^3)$ for the variance-based analysis of  policy parameters, where $H$ denotes the overall planning horizon or the number of monitoring steps in the biomanufacturing process.


Considering the substantial computational cost, an alternative way is to use sampling-based methods to enhance the estimation efficiency of SV.
\cite{castro2009polynomial} first
propose the method for the estimation of SV with permutation sampling in cooperative games,
which has been extended by \cite{song2016shapley}  to the SV analysis of simulation models.
To improve the estimation accuracy, 
\cite{mitchell2022sampling} 
show that quasi-uniform permutation points can be generated from a hypersphere and propose a permutation sampling method based on hypersphere sampling. 
However, as shown in this paper, the hypersphere sampling methods used by \cite{mitchell2022sampling} could be computationally inefficient and may 
face a challenge for cases with high-dimensional permutation space. Therefore, 
to overcome these limitations, we propose a new permutation sampling method called TFWW-VRT, which combines the hypersphere sampling method (called TFWW transformation in \cite{fang1993number}) with  two variance reduction techniques
(VRTs), i.e., randomized quasi-Monte Carlo (QMC) \citep{l2018randomized} and antithetic sampling (AS) \citep{Lemieux2009}, to generate permutations with lower computational cost and further to improve the estimation accuracy.

We summarize the main contributions of this paper as follows:
\begin{itemize} 
\item We study the SV-based sensitivity analysis  for general nonlinear pKG models and develop an efficient sensitivity estimation algorithm, which considers three types of input variables, including random factors, policy parameters and model parameters.

\item 
 For linear Gaussian pKG models, we derive the exact formula of the SV-based sensitivity analysis for random factors and design an efficient estimation method to assess the criticality of policy parameters and model parameters, which reduce the computational cost by $O(H)$ for the expectation-based SV analysis and $O(H^3)$ for the variance-based SV analysis.

\item We design an efficient permutation sampling algorithm for SV estimation for both linear and general nonlinear pKG models. Our algorithm obtains permutations efficiently from hypersphere samples with TFWW transformation, and achieves high estimation accuracy by combining with the variance reduction techniques, i.e., randomized QMC and AS methods.

\end{itemize}

The rest of this paper is organized as follows.
In Section~\ref{sec: problem_description}, we review the pKG models for biomanufacturing process and the concept of SV sensitivity analysis. In Section~\ref{sec: nonlinear_SV}, we derive the SV estimation framework for general nonlinear pKG models, while  
we present efficient SV analysis methods for linear Gaussian pKG models in Section~\ref{sec: linear_SV}.
In Section~\ref{sec: sampling}, we propose the permutation sampling method with variance reduction techniques for both general nonlinear and linear Gaussian pKG models.
We conduct empirical studies with
real data analysis
in Section~\ref{sec:empiricalStudy}, and then conclude the paper 
in
Section~\ref{sec: conclusion}. Some lengthy proofs, detailed sampling algorithm procedure, and biological kinetic equations are provided in the Appendix. The source code and data are available at \url{https://github.com/zhaojksim/SV_pKG}.


\section{Preliminaries}
\label{sec: problem_description}
In this section, we brieﬂy introduce some preliminary background and the corresponding notations used throughout the paper. We first
provide the formulation of pKG models in the context of biomanufacturing in Section~\ref{sec: hybridModeling}. Then, we review the concept of SV and the sampling-based SV estimation method in Section~\ref{sec: SV_overview}.

\subsection{Policy-Augmented Knowledge Graph}
\label{sec: hybridModeling}
Here we present the pKG models for the biomanufacturing stochastic decision process. We start with a
general nonlinear model in Section~\ref{sec: general model}, and then consider an approximated model with  linear Gaussian state transitions as well as linear rewards and control policies in Section~\ref{sec: linear Gaussian}.




\subsubsection{General Nonlinear pKG Model.}
\label{sec: general model}

A typical biomanufacturing system consists of distinct unit operations, including upstream fermentation for drug substance production and downstream purification to meet quality requirements \citep{Pauline_2013}. 
The final output metrics of the process (e.g., drug quality, productivity) are impacted by the interactions of many input-output factors, which can be divided into
critical process parameters (CPPs) and critical quality attributes (CQAs); see the definitions of CPPs/CQAs in ICH Q8(R2) \citep{guideline2009pharmaceutical}. One could think of CQAs as the ``state'' of the process, e.g., the concentrations of biomass and impurity. CPPs could be viewed as ``actions'', such as the acidity of the solution, temperature, and feed rate, that can be controlled to influence CQAs and thus optimize the output metrics.

To model the dynamic decision process of a biomanufacturing system, we use the pKG model introduced by \cite{zheng2022policy}, which is a type of finite-horizon Markov decision process (MDP) specified by the state space, action space, planning horizon, state transition probability model, reward and policy functions. At any  $t$-th time period, the state and action vectors are denoted by
$\pmb{s}_t=\left(s^1_t,...,s^n_t\right)$ and $\pmb{a}_t =\left(a^1_t,...,a^m_t\right)$, where $1 \le t \le H$ and $H$ denotes overall planning horizon or the number of monitoring steps.
The dimensions $n$ and $m$ of the state and action variables may be time-dependent, but for simplicity of notation, we keep them as constants in our presentation. The $k$-th component of the state (decision) variable is denoted by $s_t^k$ (respectively, $a^k_t$).

In the pKG model, the state transition at each time period $t$ in MDP can be modeled as 
\begin{equation}
\pmb{s}_{t+1}=f_t(\pmb{s}_t,\pmb{a}_t;\pmb{w}_t) + \pmb{e}_{t+1},
\label{eq: general transition}
\end{equation}
where $\pmb{w}_t$ denotes the model parameters of the state transition model and the residual $\pmb{e}_{t+1}$ is a random variable representing the impact of uncontrollable factors introduced in the $t$-th period. 
The structure of function $f_t$ is obtained from a mechanistic model based on domain knowledge of biological, physical, and chemical mechanisms. 
The initial state is $\pmb{s}_{1} = \pmb{s}_{0} + \pmb{e}_{1}$, where $\pmb{s}_{0}$ is the value set by the decision-makers.
We use the following example from \cite{zheng2022policy} to illustrate how bioprocess mechanisms can be incorporated into
this framework.

\begin{example}\label{example-1}
    The cell growth of Yarrowia lipolytica takes place inside a bioreactor, which requires substrate sources $S$, nitrogen $N$, subject to environmental conditions like working volume $V$, and 
 produces the target product, i.e., citrate $C$. 
During fermentation,
the feed rate $F_S$ is adjusted to control the substrate sources in the bioreactor. In this example, we can take the state variables at each period $t$ as $\pmb{s}_t = (X_{f,t}, C_t, S_t, N_t, V_t)$, where $X_f$ denotes the lipid-free
cell mass, and the action at period $t$ can be $\pmb{a}_t = F_{S, t}$. The dynamics of the state variables are modeled as kinetic equations in Appendix~\ref{secA2}, which can be conceptually written as
\begin{align}
    ds^k_t=f(\pmb{s}_t, \pmb{a}_t)dt + \sigma(s^k_t)dB_t,\nonumber
\end{align}
where $s^k_t$ for $k =1,2,\ldots, 5$ denotes the $k$-th component of $\pmb{s}_t$  in this example,  $B_t$ denotes standard Brownian noise. Following the procedure outlined in \cite{golightly2005bayesian} to convert the stochastic differential equations into discrete-time nonlinear transition models,
we can approximate the state dynamics on a small time interval $\Delta t$ by $\Delta s^k_t=f(\pmb{s}_t, \pmb{a}_t)\Delta t+\sigma(s^k_t)\sqrt{\Delta t}Z_t$, where $Z_t$ is a random variable following standard normal distribution. Let $e_{t+1}$
be a normal random variable with mean zero and standard deviation $\sigma(s_t^k)\sqrt{\Delta t}$. 
Then the state transition is approximated as
\begin{align}
s^k_{t+1}=s^k_{t}+f(\pmb{s}_t, \pmb{a}_t)\Delta t+e_{t+1}.\nonumber
\end{align}
This can be used as the transition model in the general nonlinear pKG model.  
 For instance, the transition model of $s_{t+1}^1=X_{f,t+1}$ at period $t+1$ is approximated as
\begin{align*}
    X_{f,t+1} &= X_{f,t} + \left[\mu_t - \left(\frac{F_{B, t}+F_{S, t}-V_{evap, t}}{V_t}\right)\right]X_{f,t} \Delta t + e_{t+1},
    \end{align*}
    where $\mu_t = \mu_{max, t}\big(\frac{S_t}{K_{S,t} + S_t}\cdot\frac{1}{1 + S_t/K_{iS, t}}\big)\cdot\frac{N_t}{K_{N, t} + N_t}\cdot\frac{O_t}{K_{O, t} + O_t}\cdot\frac{1}{1 + X_{f,t}/K_{ix, t}}$ is the growth rate of cells 
and 
$e_{t+1}$
is a normal random variable with mean zero and standard deviation $\sigma(X_{f,t})\sqrt{\Delta t}$. 
Note that 
the nonlinear state transition model is specified with
parameters 
$\pmb{w}_t = (\mu_{max, t}, K_{S, t}, K_{iS, t}, K_{N, t}, K_{O, t}, K_{ix, t}, O_t, F_{B, t}, V_{evap, t})$ and they are defined in Appendix ~\ref{secA2}.
\end{example}



The unknown pKG model parameters $\pmb{w} = \{\pmb{w}_t, 1 \le t < H\}$ can be estimated by using real-world data. We can see the distribution of state-action trajectory $\pmb\tau=(\pmb{s}_1,\pmb{a}_1,\pmb{s}_2,\pmb{a}_2,\ldots,\pmb{s}_{H})$ can be written as $ p(\pmb{\tau}) = p(\pmb{s}_1)\prod_{t=1}^{H-1}p(\pmb{s}_{t+1}\vert\pmb{s}_t,\pmb{a}_t)p(\pmb{a}_t\vert\pmb{s}_t)$, 
where 
$p(\pmb{s}_1)$ is the probability density function of initial state,  $p(\pmb{s}_{t+1}\vert \pmb{s}_t,\pmb{a}_t)$ denotes the conditional probability density function of $\pmb{s}_{t+1}$ given $\pmb{s}_t$ and $\pmb{a}_t$, and $p(\pmb{a}_t\vert\pmb{s}_t)$ denotes the conditional probability density function of $\pmb{a}_t$ given $\pmb{s}_t$. Given a set of real-world data 
denoted by $\mathcal{D} = \left\{\pmb\tau^{(n)}\right\}^R_{n=1}$,  we quantify the model parameter estimation uncertainty using the posterior distribution $p(\pmb{w}\vert\mathcal{D})$,
where $R$ denotes the number of samples 
and $\pmb\tau^{(n)}$ denotes the $n$-th observed trajectory for $n=1,2,\ldots,R$.
When this posterior distribution is not easy to compute, we can 
use the computational sampling approaches such as Gibbs sampling 
{(see, e.g., \citealp{Ge00})}.

\begin{figure}[!b]
\centering
\includegraphics[width=0.8\textwidth]{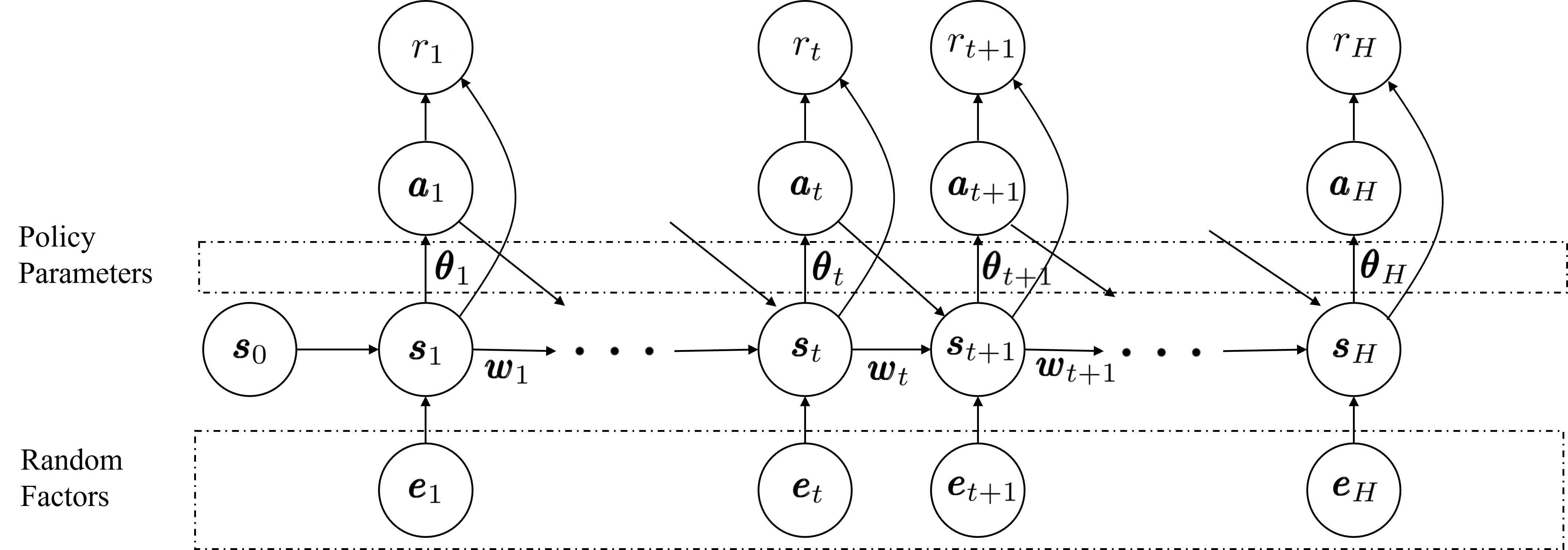}
\caption{Graphical illustration of the pKG model for biomanufacturing process. }\label{fig: pKG}
\end{figure}

Finally, we introduce the reward and policy functions in the pKG model. The process trajectory $\pmb\tau$ is evaluated in terms of revenue {depending on productivity, product CQAs,} 
{and}
 {production costs}, including the fixed cost of operating and maintaining the facility, 
and variable manufacturing costs related to raw materials, labor, quality control, and purification. In each time period, the biomanufacturing process incurs costs or rewards, which can be denoted by $r_t\left(\pmb{s}_t,\pmb{a}_t\right)$.
We model the control policy as a parametric function of the current state, i.e.,
\begin{align}
\label{eq: general_policy}
    \pmb{a}_{t} =  {h}_{t}(\pmb{s}_t, \pmb\theta_t),
\end{align}
 where $\pmb\theta_t$ denotes the policy parameters at period $t$. The 
policy $\left\{h_t\right\}^{H}_{t=1}$ is the collection of these mappings from $\pmb{s}_t$ to $\pmb{a}_t$ over the entire planning horizon. Given 
the model parameters $\pmb w$, the cumulative reward earned by policy parameters $\pmb{\theta}$ becomes, 
\begin{equation}
    J(\pmb{\theta}, \pmb\tau;\pmb w)=\sum_{t=1}^H r_t(\pmb{s}_t,\pmb{a}_t).
\end{equation}



Following the study in \cite{zheng2022policy},
we can construct the pKG model as shown in Figure~\ref{fig: pKG}. The edges connecting nodes at time $t$ to nodes at time $t+1$ (i.e., from $\pmb{s}_t$  and  $\pmb{a}_t$ to $\pmb{s}_{t+1}$) represent the process dynamics. It also consists of additional edges: 1) connecting state to action at the same time stage (i.e., from $\pmb{s}_t$ to $\pmb{a}_t$), representing the causal effect of the policy; and 2) connecting action and state to the immediate reward (i.e., from $\pmb{a}_t$ and  $\pmb{s}_t$ to $r_t$).
In the pKG model, we use Bayesian posterior distribution $p(\pmb{w}\vert\mathcal{D})$ to quantify model uncertainty 
and use the uncontrollable random factors $\pmb{e} = (\pmb{e}_1, \pmb{e}_2,\ldots,\pmb{e}_H)$ to incorporate stochastic uncertainty, i.e., variability inherent in the system. Therefore, this model can support decision-making and sensitivity analysis for biomanufacturing processes in the presence of high uncertainty. 



\subsubsection{Linear Gaussian pKG Model.}
\label{sec: linear Gaussian}

With the development of online monitoring
techniques and novel sensors facilitating real-time information collection,  CPPs/CQAs can be monitored at a faster time scale than bioprocess dynamics.
In this situation, the general nonlinear pKG model can be approximated by linear Gaussian pKG model \citep{zheng2022policy}, which can be utilized to speed up the analysis on biomanufacturing system as we illustrate in Section~\ref{sec: linear_SV}.

In the linear Gaussian pKG model, the state transition $f_t$ in Equation~(\ref{eq: general transition}) can be approximated by a linear Gaussian model, i.e.,
\begin{equation}
    \pmb{s}_{t+1} = \pmb{\mu}_{t+1}^s + \left(\pmb{\beta}_{t}^s\right)^\top(\pmb{s}_t-\pmb\mu_t^s) + \left(\pmb\beta_{t}^a\right)^\top(\pmb{a}_t-\pmb\mu_t^a) +\pmb{e}_{t+1},
    \label{eq: LGBN matrix form}
\end{equation}
where $\pmb\mu_{t}^s=(\mu_t^{1},\ldots, \mu_t^n)$ and $\pmb\mu_{t}^a=(\lambda_t^{1},\ldots, \lambda_t^m)$. The residual $\pmb{e} = (\pmb{e}_1, \pmb{e}_2,\ldots,\pmb{e}_H)$ follows multivariate Gaussian distribution.
Let $\pmb\beta_{t}^s$ denote the $n\times n$ matrix with the $\left(j,k\right) $-th element representing the linear effect of $s^j_t$ on the next state $s^k_{t+1}$. Similarly, let $\pmb\beta_t^a$ be the $m\times n$ matrix of analogous coefficients with the $\left(j,k\right)$-th element representing the linear effects of $a^j_t$ on $s^k_{t+1}$. Now the model parameters $\pmb w$ contain all entries in $\{(\pmb\beta^s_t, \pmb\beta^a_t), t=1,2,\ldots,H - 1\}$ and covariance matrix of $\pmb e$.

The biomanufacturing industry often uses 
linear reward functions (see, e.g., \citealp{martagan2016optimal,petsagkourakis2020reinforcement}), such as
\begin{equation}
\label{eq: reward_function}
r_t\left(\pmb{s}_t,\pmb{a}_t\right) = m_t + \pmb{b}^\top_t\pmb{a}_t + \pmb{c}^\top_t\pmb{s}_t,   
\end{equation}
{where the coefficients $m_t,\pmb{b}_t,\pmb{c}_t$ are nonstationary and represent both rewards and costs (that is, they can be either positive or negative)}. 

In addition, we consider the class of linear policy functions, i.e., Equation~(\ref{eq: general_policy}) becomes
\begin{equation}
   \pmb{a}_t =  {h}_{t}(\pmb{s}_t, \pmb\theta_t) = \pmb\mu^a_t + \pmb{\theta}_t^\top(\pmb{s}_t - \pmb\mu^{s}_t), \label{eq: linear policy func}
\end{equation}
where $\pmb\mu^a_t$ is the mean action value and $\pmb{\theta}_t$ is an $n\times m$ matrix of coefficients. The linear policy $h_{t}$ is thus characterized by $\pmb\theta=\left\{\pmb{\theta}_t\right\}^{H-1}_{t=1}$. The assumption of a linear policy is reasonable if the process is monitored and controlled on a sufficiently small time scale. This is because, on such a time scale, {the effect of 
state $\pmb{s}_t$ on action $\pmb{a}_t$ 
is monotonic and approximately linear. For example, the biological state of a working cell does not change quickly, so the cell's generation rate of protein/waste and uptake rate of nutrients/oxygen are roughly constant in a short time interval. Therefore, the feeding rate can be proportional to the cell density. 

\subsection{Overview of the SV Estimation}
\label{sec: SV_overview}
As we describe in Section~\ref{sec: hybridModeling}, the dynamics of pKG model are fundamentally governed by three types of input variables: the random factors $\pmb{e}$, policy parameters $\pmb{\theta}$ and model parameters $\pmb{w}$. To provide the risk- and science-based production process
understanding and guide the reliable process control,
we conduct sensitivity analysis to quantify the impact of each input variable on the output of interest, which is measured by the SV. Before we go into detail, we present the key concepts and definitions of SV that underpin our study.

In the context of sensitivity analysis, the SV of any input ${o}$ is defined as
\begin{equation}
\mbox{Sh}\left(Y\vert{o}\right)
	=\sum_{\mathcal{U}\subset \mathcal{O}/\{{o}\}}\dfrac{(\vert\mathcal{O}\vert -\vert\mathcal{U}\vert-1)! \vert\mathcal{U}\vert!} {\vert\mathcal{O}\vert!}\left[ g(\mathcal{U}\cup\left\{{o}\right\}) - g(\mathcal{U}) \right],
	\label{eq.ShapleyEffect1}
\end{equation}
where $\mathcal{O}$ denotes the set of all inputs, $\vert\mathcal{U}\vert$ is the cardinality of subset $\mathcal{U}\subset \mathcal{O}/\left\{o\right\}$ and $g(\cdot)$ is called value function associated with a set of inputs, which denotes the value of the output generated by a subset of inputs $\mathcal{U}$. The detailed forms of $g(\cdot)$ for different types of input variables are discussed in Section~\ref{sec: value_func}. We can see that $g(\mathcal{U}\cup\left\{{o}\right\}\vert\pmb{w}) - g(\mathcal{U}\vert\pmb{w})$ is the marginal contribution of input $o$ to the subsets of inputs $\mathcal{U}$ in terms of the expectation or variance of output
denoted by $g(\cdot)$. Therefore, the SV is the weighted average of the marginal contributions of $o$ to all subsets $\mathcal{U}$.

It is important to notice that as $\vert\mathcal{O}\vert$ increases, the number of subset $\mathcal{U}$ increases exponentially. To address this issue, we can rewrite Equation~(\ref{eq.ShapleyEffect1}) as
\begin{equation}
    \mbox{Sh}\left(Y\vert {o}\right)
	=\sum_{\pi \in \Pi(\mathcal{O})}\dfrac{1}{\vert \mathcal{O}\vert!}\left[ g(P_o(\pi)\cup \{o\}) - g(P_o(\pi)) \right],
	\nonumber
\label{eqn: sv_perm}
\end{equation}
where $\Pi(\mathcal{O})$ is the set of all permutations of input set $\mathcal{O}$, $P_o(\pi)$ is the set of inputs that precede $o$ in the permutation $\pi$. For instance, if the set of inputs is denoted as $\mathcal{O} = \{1, 2, 3, 4, 5\}$, one permutation of the $\mathcal{O}$ can be $\pi = \{3, 1, 2, 5, 4\}$, then $P_2(\pi) = \{3, 1\}$. \cite{castro2009polynomial} propose an approximation algorithm called ApproShapley, in which we can randomly sample $D$ permutations from $\Pi(\mathcal{O})$ and average the incremental marginal contributions $g(P_o(\pi)\cup \{o\}) - g(P_o(\pi))$ to approximate the real SV: 
\begin{equation}
    \hat{\mbox{Sh}}\left(Y\vert {o}\right)
	=\dfrac{1}{D}\sum_{d=1}^{D}\left[ g(P_o(\pi^{(d)})\cup \{o\}) - g(P_o(\pi^{(d)})) \right],
\label{eqn: sv_perm_est}
\end{equation}
where $\pi^{(d)}$ denotes the $d$-th permutation sample. We leverage such sampling-based approach to improve the estimation efficiency in the SV analysis on pKG model and further reduce the computational cost with pathway reusing in Section~\ref{sec: sv_algorithm}. 

It is also worthwhile pointing out that generating random permutations for a large number of input variables could suffer from reduced uniformity. Moreover, the direct Monte Carlo estimator in Equation~(\ref{eqn: sv_perm_est}) may not provide an accurate approximation of the SV when the computational budget $D$ is limited. To obtain high-quality samples $\{\pi^{(d)}, d = 1, \ldots, D\}$ from $\Pi(\mathcal{O})$, \mbox{\cite{mitchell2022sampling}} introduce an approach by randomly sampling points from the hypersphere and mapping these points to the permutations. They propose two hypersphere sampling methods: Box-Muller transformation (BMT) and spherical coordinate transformation (SCT), which works as follows.

Let
$U_{s}=\left\{\left(x_{1}, \ldots, x_{s}\right): x_{1}^{2}+\cdots+x_{s}^{2}=1\right\}$ denotes the set of points on $s$-dimensional hypersphere. To sample points from the uniform distribution on $U_{s}$, denoted by $\mathrm{Unif} \left(U_{s}\right)$, BMT method first generates $\pmb{z}=\left(Z_{1}, \ldots, Z_{s}\right)^{\top}$ from the $s$-dimensional standard normal distribution, and then defines $\pmb{x}=\frac{\pmb{z}}{\sqrt{\pmb{z}^{\prime}\pmb{z}}}\nonumber$, which is uniformly distributed on  $U_{s}$ \mbox{\citep{muller1959note}}. For the SCT method, if $\pmb{x}=\left(x_{1}, \ldots, x_{s}\right)^{\top} \sim \mathrm{Unif} \left(U_{s}\right)$, then it can be written as
\begin{equation}
\begin{aligned}
\label{eq: SCT}
 x_{s} &=\prod_{i=1}^{s-1} S_{i}, \quad x_{j} =\prod_{i=1}^{j-1} S_{i} A_{j}, \quad j=1, \ldots, s-1, \\
\text{ where  } S_{i}&=\sin \left(\pi \phi_{i}\right), \quad A_{i}=\cos \left(\pi \phi_{i}\right), \quad i=1, \ldots, s-2, \\
\text{ and  } S_{s-1}&=\sin \left(2 \pi \phi_{s-1}\right), \quad A_{s-1}=\cos \left(2 \pi \phi_{s-1}\right),
\end{aligned}
\end{equation}
in which the cumulative distribution function of $\phi_j$ is $F_{j}(\phi)=\frac{\pi}{B\left(\frac{1}{2}, \frac{s-j}{2}\right)} \int_{0}^{\phi}(\sin (\pi t))^{s-j-1} dt$ with $j=1, \ldots, s-1$. Therefore, SCT first samples $\phi_j$ and then generates $\pmb{x}$ based on Equation~(\ref{eq: SCT}). The detailed BMT and SCT procedures are shown in Appendices~\ref{agm: BMT} and \ref{agm: SCT}, respectively.

Note that, one drawback of the SCT procedure is that it can incur a large computational cost when the dimension of the permutation $s$ is high, which is common in biomanufacturing. BMT is more computational friendly, but it requires  $s$-dimensional normal random points, which introduces more randomness than $(\phi_1, \ldots,\phi_{s-1})$ in SCT and therefore reduces the uniformity.
To enhance the efficiency of sampling and the accuracy of SV estimation, we propose a permutation sampling algorithm that leverages a more efficient hypersphere sampling method and variance reduction techniques, as described in Section~\ref{sec: sampling}.


\section{Sensitivity Analysis on pKG Model}
\label{sec: nonlinear_SV}

In this section, we present the SV-based sensitivity analysis framework for general nonlinear pKG models, called SV-pKG. We show how to estimate SV for three types of input variables, i.e., random factors, policy parameters, and model parameters.
Notice that the form of SV depends on the type of input variables. 
Therefore,  we first introduce various value functions of SV for different input variables in Section~\ref{sec: value_func}. Then we use the SV estimation of random factors as an example to illustrate the simulation-based estimation algorithm in Section~\ref{sec: sv_algorithm}, which can reuse the calculation pathway to save the computation cost.

For the sake of clear presentation, we assume that the values of the policy parameters and the posterior distribution of model parameters are given, which means the bioprocess has been monitored by the decision-makers and learned from historical observations. 
For instance, when we estimate the SV of random factors (e.g., uncontrollable factors from the environment, and measurement noise from sensors), we fix the policy parameters and the posterior distribution of model parameters and see how the random factors impact the output to identify the critical source of variation. 



\subsection{Value Functions and SV Estimators}\label{sec: value_func}

To facilitate SV-based sensitivity analysis for different input variables in pKG model, we first need to define the corresponding value functions $g(\mathcal{U})$ associated with an input subset $\mathcal{U}$ in Equation~(\ref{eqn: sv_perm_est}), which quantify the contribution of $\mathcal{U}$ to the expectation or the variance of interested output $Y$, including future states (i.e., $Y\in\{s^{i}_t, i = 1, \ldots, n, t = 1,\ldots,H\}$) or cumulative reward (i.e., $Y=J\left(\pmb\theta,\pmb\tau;\pmb w\right)$). Based on these value functions, we subsequently develop the SV estimator for each type of input variable.
\subsubsection{Random Factors and Policy Parameters.}\label{sec: value_func_random}

We first consider the case $\mathcal{O}=\{e^{k}_t, k=1,\ldots,n ~\mbox{and} ~ t = 1,\ldots,H\}$, which includes all random factors. Given the model parameters $\pmb{w}$, we can define 
\begin{equation}
\label{eq: general_randfactor_exp}
g(\mathcal{U}\vert\pmb{w})=\mathbb{E}[Y\vert\mathcal{U};\pmb{w}],
\end{equation}
for the expectation-based (predictive) SV analysis, which represents the expected future output when the inputs in $\mathcal{U}$ are fixed as given values. For the variance-based SV analysis, let $\mbox{Var}[Y]$ denote the variance of $Y$, the value function can be defined as
\begin{equation}
g(\mathcal{U}\vert\pmb{w})=\mathbb{E}\left[\mbox{Var}\left[Y\vert\mathcal{O}/\mathcal{U};\pmb{w}\right]\right],    
\end{equation}
which is interpreted as the expected variance of the future output when the values of elements in $\mathcal{O}/\mathcal{U}$ are known. The expectation is taken on the random elements in $\mathcal{O}/\mathcal{U}$. Therefore, $g(\mathcal{U}\cup\left\{{o}\right\}\vert\pmb{w}) - g(\mathcal{U}\vert\pmb{w})$ represents the expected decrease in the variance of the output if the value of $o$ is fixed. Notice that when we define the value functions of an input variable, the policy parameters are fixed to streamline the calculation.
We omit the notation of the fixed policy parameters for simplicity. 

Then we consider the case in which the inputs in $\mathcal{O}$ include given policy parameters, i.e., $\{\theta_t^{i},i=1,\ldots,p,t=1,\ldots,H\}$. For the predictive SV analysis, to quantify the contribution of the policy parameters to the expected future output, 
we can define 
\begin{equation}
\label{eq: value_theta}
g(\mathcal{U}\vert\pmb{w})=\mathbb{E}[Y\vert\pmb{\theta}_{\mathcal{O}/\mathcal{U}}=0; \pmb{w}],
\end{equation}
where $\pmb{\theta}_{\mathcal{O}/\mathcal{U}}=0$ means the policy parameters $\theta_t^{i}$ that are not in $\mathcal{U}$ are set to be zero, i.e., the baseline value or uninformative value \citep{covert2021explaining, ancona2019explaining, lundberg2017unified1}.
Similarly, for the variance-based SV analysis, we define 
\begin{equation}
\label{eq: value_theta_var}
    g(\mathcal{U}\vert\pmb{w})=\mbox{Var}[Y \vert\pmb{\theta}_{\mathcal{O}/\mathcal{U}}=0;\pmb{w}],
\end{equation}
which measures the variance of the output when the policy parameter elements not in $\mathcal{U}$ are set to be zero. Notice that the policy parameters are deterministic in our setting, so there is no outer expectation taken on policy parameters.

So far we assume that the model parameters are given when we define the value functions for random factors and policy parameters. To incorporate the model uncertainty in the SV estimation, we first sample $Q$ sets of model parameters from $p(\pmb{w}\vert\mathcal{D})$. Then we sample $D$ permutations for given model parameter $\pmb{w}^{(q)}$ with $q=1,\ldots,Q$. 
For each sampled permutation, as the expectation and the variance in the value functions cannot be obtained directly for general nonlinear pKG models, we need to simulate the value function according to the pKG model, which captures the mechanism and the stochasticity in the biomanufacturing process. 
For the SV of input $o$, denoted by $\mbox{Sh}\left(Y\vert {o}\right)$, we adopt the following estimator:
\begin{equation}   \hat{\mbox{Sh}}\left(Y\vert {o}\right) =\dfrac{1}{QD}\sum_{q=1}^{Q}\sum_{d=1}^{D}\left[ \hat{g}(P_o(\pi^{(d)})\cup \{o\}\vert \pmb{w}^{(q)}) - \hat{g}(P_o(\pi^{(d)})\vert \pmb{w}^{(q)}) \right],
\label{eq: svestimator}
\end{equation}
where $\hat{g}(P_o(\pi^{(d)})\vert \pmb{w}^{(q)})$ is the simulation estimation of $g(P_o(\pi^{(d)})\vert \pmb{w}^{(q)})$ and $\pi^{(d)}$ is $d$-th permutation sample. If $o$ is the first element of $\pi^{(d)}$, then $P_o(\pi^{(d)}) = \emptyset$ is the empty set.
We finally obtain the estimation of the SV for each input $o$ by averaging their marginal contributions under different model parameters and permutations. Some properties of the Shapley value estimator, i.e., Equation~(\ref{eq: svestimator}), are shown in Proposition~\ref{prop: properties} and their proofs can be found in Appendix~\ref{subsec: proof_properties}, which are obtained from \cite{castro2009polynomial} by considering model uncertainty:

\begin{proposition}
\label{prop: properties}
The SV estimator in Equation (3) satisfies the following properties:
\begin{roster}
    \item Suppose that the value function $g$ can be estimated exactly, it is an unbiased and consistent estimator for
    $\mbox{Sh}\left(Y\vert {o}\right)$ and its variance is given by
    $\dfrac{\sigma^{2}}{QD}$, where
    \begin{equation}
    \begin{aligned}
    \sigma^{2} =  \dfrac{\sum_{\pi \in\Pi(\mathcal{O})}\mathbb{E}_{\pmb{w}}\{\left[g(P_o(\pi)\cup \{o\}\vert \pmb{w}) - g(P_o(\pi)\vert\pmb{w}) - \mbox{Sh}\left(Y\vert {o}\right)\right]^2\}}{\vert \mathcal{O}\vert!}.
    \nonumber
    \end{aligned}
    \end{equation}
    \item It is efficient in the allocation, i.e.,
    \begin{equation}
    \sum_{o\in \mathcal{O}}\hat{Sh}\left(Y\vert {o}\right)= \dfrac{1}{QD}\sum_{q=1}^{Q}\sum_{d=1}^{D}\left[ \hat{g}(\mathcal{O}\vert \pmb{w}^{(q)}) - \hat{g}(\emptyset\vert \pmb{w}^{(q)}) \right],
    \nonumber
    \end{equation}
    which means the sum of the estimated SVs of inputs in $\mathcal{O}$ equals to the estimated marginal value generated by $\mathcal{O}$.
\end{roster}
\end{proposition}

\subsubsection{Model Parameters.}\label{sec: value_func_model}

Finally, we consider the scenario in which model parameters are employed as input variables, i.e., $\mathcal{O} = \{w^{i}_t, i=1,\ldots,\mathcal{I}, t=1,\ldots,H\}$, where for simplicity we assume that the number of model parameters at each period $t$ is $\mathcal{I}$ and denote the $i$th model parameter at period $t$ as $w^{i}_t$. The value functions of model parameters can be specified in a similar manner as those of random factors, except that they omit the conditioned model parameters as they are now considered as inputs. In particular, for the predictive SV analysis, the value function becomes
\begin{equation}
    \label{eq: value_model_predictive}
g(\mathcal{U})=\mathbb{E}[Y\vert\mathcal{U}],
\end{equation}
which means the expected output when the model parameters in $\mathcal{U}$ are fixed as given values. 
Given finite real-world observations, the uncertainty in the model parameters characterizing our limited knowledge
on the probabilistic interdependence of integrated
bioprocess. To study the impact of model uncertainty using variance-based SV analysis, we define the value function as
\begin{equation}
g(\mathcal{U})=\mathbb{E}\left[\mbox{Var}\left[Y\vert\mathcal{O}/\mathcal{U}\right]\right],
\end{equation}
which means the expectation of remaining  variance when all other model parameters in $\mathcal{O}/\mathcal{U}$ are fixed.
The resulting estimator for the SV of one model parameter $o\in\mathcal{O}$ is
\begin{equation*}
\hat{\mbox{Sh}}\left(Y\vert {o}\right) =\dfrac{1}{D}\sum_{d=1}^{D}\left[ \hat{g}(P_o(\pi^{(d)})\cup \{o\}) - \hat{g}(P_o(\pi^{(d)})) \right].
\end{equation*}
The SV of model parameters can provide
the comprehensive and interpretable understanding on how
model uncertainty impacts the process risk and
identify those parameters contributing the most to the
estimation uncertainty. By ranking the SV, we can conduct efficient data collection to reduce the
impact of model estimation uncertainty.
\subsection{SV Estimation Algorithm}
\label{sec: sv_algorithm}

\begin{figure}[!b]
\centering
\includegraphics[width=0.7\textwidth]{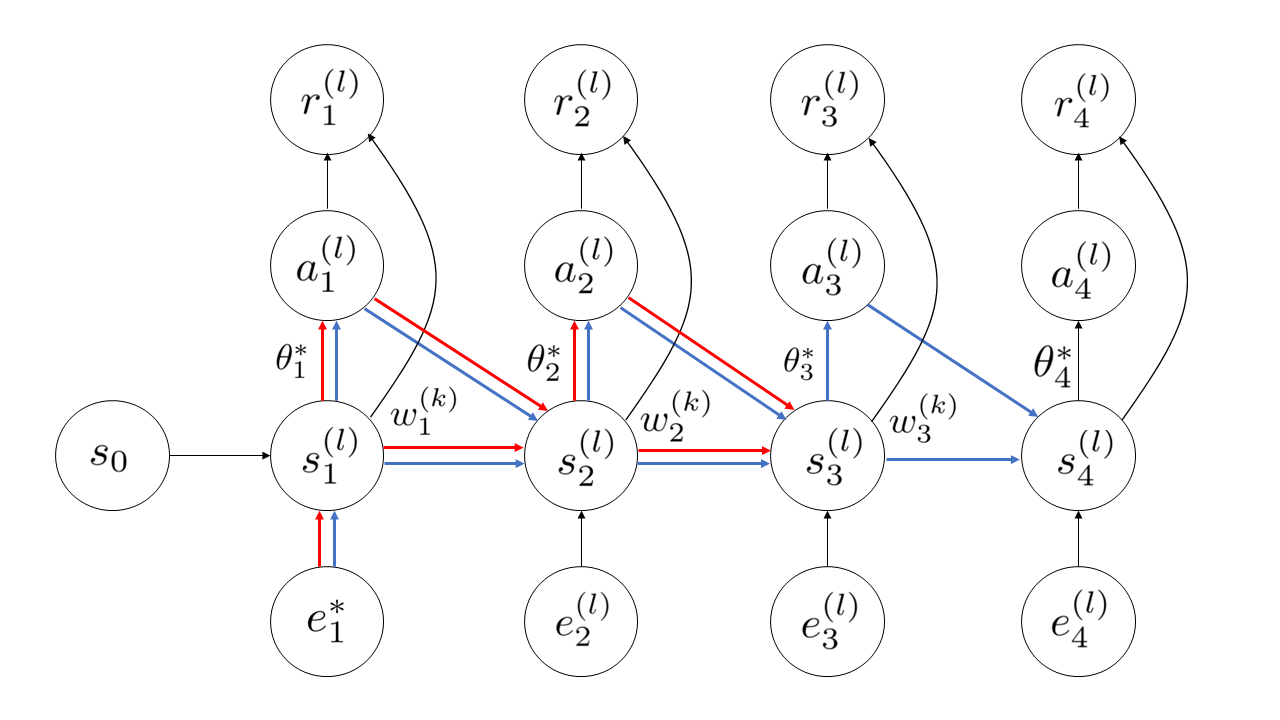}
\caption{The reused pathway for the simulation of value function.}\label{propagation}
\end{figure}

In this section, we introduce the SV estimation algorithm for general nonlinear pKG models and implementation details for general nonlinear pKG models. The procedure for SV estimation is shown in Algorithm~\ref{agm: svgeneral}. 
We take the predictive SV analysis of random factors as an example and consider the states at each period as the outputs of interest, i.e., $\mathcal{O} =\{e_t^k,1\le k\le n, 1\le t\le H\}$, $g(\mathcal{U}\vert\pmb{w})=\mathbb{E}[Y\vert\mathcal{U};\pmb{w}]$ in Equation~(\ref{eq: general_randfactor_exp}), and $Y\in\{\pmb{s}_t, t = 1,\ldots,H\}$. Specifically, lines 3-7 in Algorithm~\ref{agm: svgeneral} sample $Q$ model parameters and $D$ random permutations, where the permutation sampling is based on the Algorithm~\ref{algosampling}, introduced  later in Section~\ref{sec: sampling}. Lines 8-24 first incrementally estimate the marginal contribution of each input $o$ to $\pmb{s}_t$, $1\le t\le H$, for given sample $\pmb{w}^{(q)}$ and $\pi^{(d)}$, then accumulate it in $\hat{Sh}^{\pmb{s}_t}_o$, $1\le t\le H$. Lines 28-29 average the cumulative marginal contribution as in Equation~(\ref{eq: svestimator}).

\SetInd{0.5em}{1em}
\SetNlSkip{1em}
\begin{algorithm}[!t]
\linespread{1.2}\selectfont
Input: the number of samples for model parameters $Q$, the number of permutations $D$, the initial state value $\pmb{s}_0$, the policy parameters $\pmb{\theta}$, the distribution of model parameters $p(\pmb{w}\vert \mathcal{D})$, the set of inputs $\mathcal{O} =\{e_t^k,1\le k\le n, 1\le t\le H\}$,
$\hat{Sh}^{\pmb{s}_t}_o  = \pmb{0} \text{ for } o \in \mathcal{O}$ and $1\le t\le H$.
\\
Generate the samples of the marginal contribution and calculate the sum of samples:\\
\For{$q = 1,2,\ldots,Q$}{
Sample model parameters from the posterior distribution:\\ 
$\pmb{w}^{(q)}\sim p(\pmb{w}\vert\mathcal{D})$;\\
\For{$d=1,2,\ldots,D$}{
Sample the random permutation $\pi^{(d)}$ from $\Pi(\mathcal{O})$ using Algorithm~\ref{algosampling};\\
\For{$i=0,1,\ldots,\vert \mathcal{O}\vert$}{
\eIf{$i = 0$}{$\mathcal{U} = \emptyset$, $\hat{g}_{prev}^{\pmb{s}_t} = 0$, $1\le t\le H$;}{$\mathcal{U} = \mathcal{U}\cup\{\pi^{(d)}(i)\}$, where $\pi^{(d)}(i)$ is the $i$th element of $\pi^{(d)}$;
\\ $\hat{g}^{\pmb{s}_t}_{prev}=\hat{g}^{\pmb{s}_t}$, $1\le t\le H$;}
\tikzmark{s1}\For{$l=1,\ldots, L$}{
Sample $\pmb{e}^{(l)} = (\pmb{e}^{(l)}_1, \pmb{e}^{(l)}_2, \ldots, \pmb{e}^{(l)}_H)$\Comment*[r]{Reusing pathway}
Set the random factors in $\mathcal{U}$ to be predefined value;\\
$\pmb{s}^{(l)}_{1} = \pmb{s}_0 + \pmb{e}^{(l)}_1$;\\
\For{$t=1,\ldots, H-1$}{
$\pmb{a}^{(l)}_t = {h}_{t}(\pmb{s}^{(l)}_{t};\pmb\theta_t)$,
$\pmb{s}^{(l)}_{t+1} = f_t(\pmb{s}^{(l)}_{t}, \pmb{a}^{(l)}_t;\pmb{w}^{(q)}_t) + \pmb{e}^{(l)}_{t+1}$;
}}
$\hat{g}^{\pmb{s}_t} = \sum_{l=1}^{L}(\pmb{s}^{(l)}_t)/L$, $1\le t\le H$;   \hspace{4.2cm}\tikzmark{e1}
\drawCodeBox{s1}{e1}\\
\vspace{0.3em}
$\hat{Sh}^{\pmb{s}_t}_{\pi^{(d)}(i)} = \hat{Sh}^{\pmb{s}_t}_{\pi^{(d)}(i)}+\hat{g}^{\pmb{s}_t}-\hat{g}^{\pmb{s}_t}_{prev}$, $1\le t\le H$;}}}
Calculate the estimation of the SV:\\
$\hat{Sh}^{\pmb{s}_t}_o = \hat{Sh}^{\pmb{s}_t}_o/(QD) \text{ for } o \in \mathcal{O}$, $1\le t\le H$.
\caption{Predictive SV Analysis on Random Factors of pKG
}\label{agm: svgeneral}
\end{algorithm}	
In practice, there may be multiple outputs of interest, which may be related to the CQAs of the biomanufacturing process at different positions and different periods. If we simulate the value function $g(\cdot\vert\pmb{w})$ for different outputs separately, i.e., the brute force computation algorithm, the algorithm would be computationally expensive as the time horizon increases. We can utilize the hierarchical temporal structure of the pKG models, which means we can generate and record all CQAs simultaneously in one sample trajectory. For the example shown in Figure~\ref{propagation}, the overall time horizon is $H = 4$ and there are one CPP and one CQA in each period. If we are interested in the contribution of random factor $e_1$ to the expectation of $s_3$ and $s_4$ at $e_1=e_1^*$ given policy $\pmb\theta^*$, we need to sample $e_2$ and $e_3$ and obtain the $l$th sample $(e_2^{(l)}, e_3^{(l)})$. Then we can calculate $s_3^{(l)}$ by forward propagation along the pathway from $e^*_1$ to $s_3$ in the pKG model, which is highlighted in red. Now $s_4^{(l)}$ can be calculated based on $s_3^{(l)}$ and sample $e_4^{(l)}$, whose propagation pathway is highlighted in blue. Therefore, the algorithm reuses the calculation of $s_3^{(l)}$ due to the shared trajectory path, which can avoid the unnecessary computational burden. 
If the time horizon $H$ is large, this algorithm can reduce the computational cost by a
factor of $O(H)$ compared with the brute force algorithm
that does not reuse calculations. The detailed procedure for reusing calculation pathway is shown in Algorithm~\ref{agm: svgeneral}, lines 15-23.

We conclude the algorithm section with a discussion on how to determine a proper sample size (i.e., $QD$ for the random factors and policy parameters, $D$ for the model parameters) in order to bound the probability that the estimated SV deviates at most $\epsilon$ from the true exact SV, where $\epsilon$ denotes a positive constant. For illustration, we take the SV estimation of random factors and policy parameters as an example and the decision rule for the sample size is as follows:
\begin{proposition}
\label{prop: sample_size}
Let ${Sh}\left(Y\vert {o}\right)$ and $\hat{Sh}\left(Y\vert {o}\right)$ be the exact and the estimated SV of an input, respectively. To ensure $\Pr(\vert\hat{Sh}\left(Y\vert {o}\right)-{Sh}\left(Y\vert {o}\right)\vert\ge \epsilon) \le \delta$,
 \begin{roster}
   \item if the variance $\sigma^2$ of the marginal contributions for the input $o$ is known, we can set 
   \begin{equation}
   \label{eq: sample_size_sigma}
       QD \ge \lceil \dfrac{\sigma^2}{\delta\epsilon^2} \rceil,
   \end{equation}
   where $\lceil x \rceil$ denotes the smallest integer that is not less than $x$;
   \item if the range $r$ of the marginal contributions for the input $o$ is known, we can set 
   \begin{equation*}
   QD \ge \lceil \dfrac{\ln(2/\delta)r^2}{2\epsilon^2} \rceil.  
   \end{equation*}
  \end{roster}
    \end{proposition}
The detailed proof of Proposition~\ref{prop: sample_size} is shown in Appendix~\ref{subsecA: proof_sample_size}. Proposition~\ref{prop: sample_size} provides guidance on determining the required sample size in SV estimation.  One concern is that there are multiple inputs and outputs in the SV analysis of the pKG model. In practice, we can weight the sample size of each input by the variance of the marginal contributions and calculate the weighted average sample size  for a given output. Then, we take the maximum sample size among the different outputs as the required sample size.



\section{Sensitivity Analysis for Linear Gaussian pKG}
\label{sec: linear_SV}

In this section, we focus on the SV analysis for linear Gaussian pKG models. In modern biomanufacturing, the production process could be monitored on a faster
time scale than the evolution of bioprocess dynamics. As the number of monitoring steps $H$ increases, the SV estimation of input variables can be time-consuming. To estimate the SV efficiently, we use 
the linear Gaussian pKG model to approximate the behavior of such heavily instrumented biomanufacturing processes. In Section~\ref{sec: random_SV}, we derive the analytical SV formula for random factors. As the output of interest is not linear with the policy parameters and model parameters, there is a lack of exact SV formulas for policy parameters and model parameters, and the naive implementation of the simulation-based approach is time-consuming when $H$ is large, we propose efficient estimation algorithms for the SVs of policy parameters and model parameters in Section~\ref{sec: policyapproxSV} and Section~\ref{sec: linear_modelsv}, respectively. By utilizing the structural property of the linear Gaussian pKG model, we can conduct the SV analysis on the policy parameters and model parameters with less computational cost.  

\subsection{SV Formula of Random Factors}
\label{sec: random_SV}


In this section, we focus on the SV analysis on random factors, i.e., $\mathcal{O} = \{e_t^k,k=1,\ldots,n, t=1,\ldots,H\}$, for linear Gaussian pKG models. Notice that the random factors correspond to the residual nodes in pKG models. Therefore, given a set of model parameters, the SVs of random factors can be calculated exactly under the linear Gaussian assumption.

We first consider the predictive SV analysis on random factors. According to the transition model in Equations~(\ref{eq: LGBN matrix form}) and~(\ref{eq: linear policy func}), we have
\begin{equation}
\label{eq: state_transition}
\pmb{s}_{t+1}=\pmb{\mu}_{t+1}^s +\mathbf{R}_{1,t}(\pmb{s}_{0} - \pmb\mu^{s}_{1}) +\left[\sum_{i=1}^{t}\mathbf{R}_{i,t}\pmb{e}_i + \pmb{e}_{t+1}\right],
\end{equation}
where $\mathbf{R}_{i,t} =\prod_{j=t}^i\left[\left(\pmb\beta_{j}^s\right)^\top + \left(\pmb\beta_{j}^a\right)^\top\pmb\theta_{j}^\top\right]$ represents the product of the pathway coefficients from period $i$ to period $t+1$ and $\prod\limits_{j=t}^{i}$ denotes the product of coefficients from index $t$ to index $i$. If $t = i - 1$, $\mathbf{R}_{i,t} = \mathbf{I}$, the identity matrix; If $t < i-1$, $\mathbf{R}_{i,t} = \mathbf{0}$, the zero matrix.
According to Equations~(\ref{eq: state_transition}) and~(\ref{eq: reward_function}),
\begin{align*}
\mathbb{E}\left(\pmb{s}_{t+1}\vert e_h^k, \pmb{\theta}; \pmb{w}\right)
&= \mathbb{E}\left\{ \pmb{\mu}_{t+1}^s +\mathbf{R}_{1,t}(\pmb{s}_{0} - \pmb\mu^{s}_{1}) +\left[\sum_{i=1}^{t}\mathbf{R}_{i,t}\pmb{e}_i + \pmb{e}_{t+1}\right]\vert e_h^k,\pmb{\theta};  \pmb{w}\right\}\\
&=\pmb{\mu}_{t+1}^s +\mathbf{R}_{1,t}(\pmb{s}_{0} - \pmb\mu^{s}_{1})+\mathbf{R}_{h,t}e_h^k\mathrm{1}_k,\\
\mathbb{E}\left(r_{t+1}\vert e_h^k,\pmb{\theta};\pmb{w}\right)
&=
\mathbb{E}\Biggl\{\pmb{b}_{t+1}^\top\pmb{a}_{t+1}+\pmb{c}_{t+1}^\top \pmb{s}_{t+1}+m_{t+1}\vert e_h^k, \pmb{\theta};  \pmb{w}\Biggl\}\\
&=(\pmb{b}_{t+1}^\top\pmb{\theta}_{t+1}^\top+\pmb{c}_{t+1}^\top) \mathbf{R}_{h,t}e_h^k\mathrm{1}_k+M_{t+1},
\end{align*}
where $e_h^k$ is the $k$-th random factor in period $h$, $\mathrm{1}_k$ is the standard basis column vector whose components are all zero except the $k$-th component, which is equal to 1, and  $M_{t+1} = m_{t+1} + \pmb{b}_{t+1}^\top \pmb\mu^a_{t+1}+\pmb{c}_{t+1}^\top\pmb{\mu}_{t+1}^s+(\pmb{b}_{t+1}^\top\pmb{\theta}_{t+1}^\top+\pmb{c}_{t+1}^\top)\mathbf{R}_{1,t}(\pmb{s}_{0} - \pmb\mu^{s}_{1})$. Theorem~\ref{thm: predictive SV for random factors} calculates the SV, with detailed derivations deferred to Appendix~\ref{subsecA: proof_them1}.
\begin{theorem}
\label{thm: predictive SV for random factors}
For the predictive sensitivity analysis of the random factor $e^k_h$, the SV is given by 
\begin{align*}
\mbox{Sh}\left(\pmb{s}_{t+1}\vert {o};\pmb{\theta}\right) &=\mathbb{E}_{\pmb{w}}\left\{\mathbf{R}_{h,t}e_h^k\mathrm{1}_k\right\},\\
\mbox{Sh}\left(\sum_{t=0}^{H-1} r_{t+1}\vert {o};\pmb{\theta}\right)&=\mathbb{E}_{\pmb{w}}\left\{\sum_{t=0}^{H-1}(\pmb{b}_{t+1}^{\top}\pmb{\theta}_{t+1}^\top+\pmb{c}_{t+1}^\top)\mathbf{R}_{h,t}e_h^k\mathrm{1}_k\right\}.
\end{align*}
\end{theorem}
Theorem~\ref{thm: predictive SV for random factors} provides the exact formula of SV in the predictive SV analysis of random factors for linear Gaussian pKG models. Note that $\mathbf{R}_{h,t}$ is the pathway coefficients from period $h$ to period $t+1$, and we can see the SV of $e_h^k$ is the propagated impact of $e_h^k$ on each output.

We now consider the variance-based SV of random factors. 
Let $\pmb{\alpha}_{t+1}=\pmb{b}_{t+1}^\top\pmb{\theta}_{t+1}^\top+\pmb{c}_{t+1}^\top$, $\pmb{X}_{t+1} = (\pmb{e}_1^{\top},\pmb{e}_2^{\top},...,  \pmb{e}_{t+1}^{\top})^{\top}$ and $\mathbf{R}_{t+1} = \left( \mathbf{R}_{1,t},...,\mathbf{R}_{t,t}, \mathbf{I}\right)$. The covariance matrix of $\pmb{X}_{t+1}$ is $\mathbf{V}_{t+1}$ and the covariance matrix between $\pmb{X}_{t_1+1}$ and $\pmb{X}_{t_2+1}$ is $\mathbf{V}_{t_1+1, t_2+1}$. $\mathbf{1}_{t+1}^{l}$ is the matrix whose shape is the same as $\mathbf{V}_{t+1}$ and the entries in $l$th column are 1 and other entries in other columns are 0, and $\mathbf{1}_{t_1+1, t_2+1}^{l}$ is the matrix whose shape is the same as $\mathbf{V}_{t_1+1, t_2+1}$ and the entries in $l$th column are 1 and other entries in other columns are 0. If the random factor $e^k_h$ is the $l$th entry of $\pmb{X}_{t+1}$ and is at period $h$, Theorem~\ref{thm: variance-based SV for random factors} calculates the variance-based SV for random factors and the detailed derivation can be seen in Appendix~\ref{subsecA: proof_them2}.
\begin{theorem}
\label{thm: variance-based SV for random factors}
Let $\odot$ denote component-wise multiplication. For the variance-based sensitivity analysis of the random factor $e^k_h$, the SV is given by 
\begin{align*}
\label{thm: var shapley for random factors}
\mbox{Sh}\left(\pmb{s}_{t+1}\vert {o};\pmb{\theta}\right) &=\mathbb{E}_{\pmb{w}}\left\{\mathbf{R}_{t+1}\left[\mathbf{V}_{t+1} \odot \left[\frac{1}{2}(\mathbf{1}_{t+1}^{l}+\mathbf{1}_{t+1}^{l\top})\right]\right]\mathbf{R}_{t+1}^{\top}\right\},\\
\mbox{Sh}\left(\sum_{t=0}^{H-1} r_{t+1}\vert {o};\pmb{\theta}\right)&=\mathbb{E}_{\pmb{w}}\left\{\mbox{Sh}\left(\sum_{t=0}^{H-1} r_{t+1}\vert{o},\pmb{\theta}; \pmb{w}\right)\right\},
\end{align*}
where 
\begin{align*}
&\mbox{Sh}\left(\sum_{t=0}^{H-1} r_{t+1}\vert{o},\pmb{\theta};\pmb{w}\right)=\sum_{t=h}^{H-1}\pmb{\alpha}_{t+1}\mathbf{R}_{t+1}\left[\mathbf{V}_{t+1} \odot \left[\frac{1}{2}(\mathbf{1}_{t+1}^{l}+\mathbf{1}_{t+1}^{l\top})\right]\right]\mathbf{R}_{t+1}^{\top}\pmb{\alpha}_{t+1}^{\top}\\
&+\sum_{t_2=h}^{H-1}\sum_{t_1=0}^{h-1}\pmb{\alpha}_{t_1+1}\mathbf{R}_{t_1+1}\left[\mathbf{V}_{t_1+1, t_2+1}\odot(\dfrac{1}{2}\mathbf{1}_{t_1+1, t_2+1}^{l})\right]\mathbf{R}_{t_2+1}^{\top}\pmb{\alpha}_{t_2+1}^{\top}\\
&+\sum_{t_2=2}^{H-1}\sum_{t_1=h}^{t_2-1}\pmb{\alpha}_{t_1+1}\mathbf{R}_{t_1+1}\left[\mathbf{V}_{t_1+1, t_2+1}\odot\left[\dfrac{1}{2}(\mathbf{1}_{t_1+1, t_2+1}^{l}+\mathbf{1}_{t_2+1, t_1+1}^{l\top})\right]\right]\mathbf{R}_{t_2+1}^{\top}\pmb{\alpha}_{t_2+1}^{\top}.
\end{align*}
\end{theorem}
In Theorem~\ref{thm: variance-based SV for random factors}, $\pmb{\alpha}_{t+1}$, $\mathbf{R}_{t+1}$ and $\mathbf{V}_{t+1}$ can be calculated once and stored for reuse in the computation of variance-based SV. From Theorem~\ref{thm: variance-based SV for random factors}, we can see the SV of a random factor depends on both the variation of itself and the interaction of other random factors on the pathway between this random factor and the output. The biomanufacturing process may amplify such variation by the transition mode. Therefore, we can reduce the variance of the output by carefully monitoring the most fluctuating random factors, mitigating its impact by adjusting the policy parameters and reducing the positive correlations between the random factors.


\subsection{SV Estimation for Policy Parameters}
\label{sec: policyapproxSV}
In this section, we conduct the SV analysis on policy parameters and model parameters for linear Gaussian pKG models. As there is no exact formula for the SV of policy parameters, we need to approximate the SV by permutation sampling. Note that, under the linear Gaussian assumption, the value function can be derived exactly. We show how to efficiently calculate the value function in the SV estimation of the policy parameters and model parameters.

To quantify the contribution of the policy parameters to the expected future state $\pmb{s}_{t+1}$, the value function defined by Equation~(\ref{eq: value_theta}) in Section~\ref{sec: nonlinear_SV} becomes 
\begin{equation}
\begin{aligned}
g(\mathcal{U}\vert\pmb{w})&=\mathbb{E}[\pmb{s}_{t+1}\vert \pmb{\theta}_{\mathcal{O}/\mathcal{U}}=0;\pmb{w}]\\
&=\mathbb{E}\left\{\pmb{\mu}_{t+1}^s +\mathbf{R}_{1,t}(\pmb{s}_{0} - \pmb\mu^{s}_{1}) +\left[\sum_{i=1}^{t}\mathbf{R}_{i,t}\pmb{e}_i + \pmb{e}_{t+1}\right]\vert\pmb{\theta}_{\mathcal{O}/\mathcal{U}}=0; \pmb{w}\right\}\\
&=\pmb{\mu}_{t+1}^s +\mathbf{R}_{1,t}(\pmb{s}_{0} - \pmb\mu^{s}_{1})\vert \pmb{\theta}_{\mathcal{O}/\mathcal{U}}=0;\pmb{w}.
\nonumber
\end{aligned}
\end{equation}
For the cumulative reward, the value function is 
\begin{equation}
\begin{aligned}
\label{eq: linear_value_policy_predictive}
g(\mathcal{U}\vert\pmb{w})&=\mathbb{E}\left[\sum_{t=0}^{H-1}r_{t+1}\vert\pmb{\theta}_{\mathcal{O}/\mathcal{U}}=0;\pmb{w}\right]\\
&=\sum_{t=0}^{H-1}\left\{m_{t+1} + \pmb{b}_{t+1}^\top \pmb\mu^a_{t+1}+\pmb{c}_{t+1}^\top\pmb\mu^s_{t+1}+\pmb\alpha_{t+1}\left[\mathbf{R}_{1,t}(\pmb{s}_{0} - \pmb\mu^{s}_{1})\right]\vert \pmb{\theta}_{\mathcal{O}/\mathcal{U}}=0;\pmb{w}\right\}.
\end{aligned}
\end{equation}
The calculation of the value function requires the calculation of $\mathbf{R}_{1,t}$, $t = 1,2,\ldots,H-1$. As
\begin{equation}
\begin{aligned}
\mathbf{R}_{1,t} &=\prod_{j=t}^{1}\left[\left(\pmb\beta_{j}^s\right)^\top + \left(\pmb\beta_{j}^a\right)^\top\pmb\theta_j^\top\right]\\
&=\left[\left(\pmb\beta_{t}^s\right)^\top + \left(\pmb\beta_{t}^a\right)^\top\pmb\theta_{t}^\top\right]\mathbf{R}_{1,t-1},
\end{aligned}
\label{eq: nested_propogation}
\end{equation}
 we can recursively calculate the $\mathbf{R}_{1,t}$, $t = 1,2,\ldots,H-1$. Therefore, predictive SV analysis using Equation~(\ref{eq: nested_propogation}) can reduce the computational burden by reusing calculations. The detailed procedure is described as  Algorithm~\ref{agm: sv_linear_predictive} in Appendix~\ref{sec: agm_policy}. Proposition~\ref{prop: time complexity analysis for predictive SV} shows the current time complexity of the SV calculation. 
\begin{proposition}
\label{prop: time complexity analysis for predictive SV}
Given the sample size of the model parameters $Q$ and permutations $D$, the cost to compute the approximate predictive SV of policy parameters with value function in Equation~(\ref{eq: value_theta}) for linear Gaussian pKG using Equation~(\ref{eq: nested_propogation}) is $O\left(QDH^2(n^4m+n^3m^2)\right)$. For the brute force algorithm without reusing calculations, the cost is $O\left(QDH^3(n^4m+n^3m^2)\right)$.
\end{proposition}
Compared to the brute force algorithm that calculates the value function directly, without reusing calculation, the recursive computation can save the computational cost by a factor of $O(H)$. The detailed proof can be found in Appendix~\ref{subsecA: proof_prop5}.

When we want to quantify the contribution of the policy parameters to the variances of the states and the cumulative reward, the inputs in $\mathcal{O}$ include policy parameters. As defined in Equation~(\ref{eq: value_theta_var}), the value function is
\begin{equation}
\begin{aligned}
\label{eq: linear_value_theta_var}
g(\mathcal{U}\vert\pmb{w})&=\mbox{Var}[\sum_{t=0}^{H-1} r_{t+1}\vert \pmb{\theta}_{\mathcal{O}/\mathcal{U}}=0; \pmb{w}]\\
&=\sum_{t_1 = 0}^{H-1}\sum_{t_2 = 0}^{H-1}\mbox{Cov}(\pmb{s}_{t_1+1}, \pmb{s}_{t_2+1}\vert \pmb{\theta}_{\mathcal{O}/\mathcal{U}}=0; \pmb{w}).
\end{aligned}
\end{equation} 
To calculate $g(\mathcal{U}\vert\pmb{w})$, we need to calculate the covariance matrix between $\pmb{s}_{t_1+1}$ and $\pmb{s}_{t_2+1}$:
\begin{equation}
\label{eq: linear_policy_cov}
\begin{aligned}
\mbox{Cov}(\pmb{s}_{t_1+1}, \pmb{s}_{t_2+1})&=\sum_{i=1}^{t_1}\sum_{j=1}^{t_2}\mathbf{R}_{i, t_1}\mbox{Cov}(\pmb{e}_i,\pmb{e}_j)\mathbf{R}_{j, t_2}^{\top}+
\sum_{i=1}^{t_1}\mathbf{R}_{i, t_1}\mbox{Cov}(\pmb{e}_i,\pmb{e}_{t_2+1})\\
&\quad+\sum_{i=1}^{t_2}\mbox{Cov}(\pmb{e}_{t_1+1},\pmb{e}_i)\mathbf{R}_{i, t_2}^{\top}+\mbox{Cov}(\pmb{e}_{t_1+1}, \pmb{e}_{t_2+1}),\\
\mbox{Cov}(\pmb{s}_{t_1+1}, \pmb{s}_{t_2})&=\sum_{i=1}^{t_1}\sum_{j=1}^{t_2-1}\mathbf{R}_{i, t_1}\mbox{Cov}(\pmb{e}_i,\pmb{e}_j)\mathbf{R}_{j, t_2-1}^{\top}+
\sum_{i=1}^{t_1}\mathbf{R}_{i, t_1}\mbox{Cov}(\pmb{e}_i,\pmb{e}_{t_2})\\
&\quad+\sum_{i=1}^{t_2-1}\mbox{Cov}(\pmb{e}_{t_1+1},\pmb{e}_i)\mathbf{R}_{i, t_2-1}^{\top}+\mbox{Cov}(\pmb{e}_{t_1+1}, \pmb{e}_{t_2}),\\
\mbox{Cov}(\pmb{s}_{t_1}, \pmb{s}_{t_2+1})&=\sum_{i=1}^{t_1-1}\sum_{j=1}^{t_2}\mathbf{R}_{i, t_1-1}\mbox{Cov}(\pmb{e}_i,\pmb{e}_j)\mathbf{R}_{j, t_2}^{\top}
+\sum_{i=1}^{t_1-1}\mathbf{R}_{i, t_1-1}\mbox{Cov}(\pmb{e}_i,\pmb{e}_{t_2+1})\\
&\quad+\sum_{i=1}^{t_2}\mbox{Cov}(\pmb{e}_{t_1},\pmb{e}_{i})\mathbf{R}_{i, t_2}^{\top}+\mbox{Cov}(\pmb{e}_{t_1}, \pmb{e}_{t_2+1}).
\end{aligned}
\end{equation}

For the first part of the $\mbox{Cov}(\pmb{s}_{t_1+1}, \pmb{s}_{t_2+1})$, we have
\begin{align*}
\sum_{i=1}^{t_1}\sum_{j=1}^{t_2}\mathbf{R}_{i, t_1}\mbox{Cov}(\pmb{e}_i,\pmb{e}_j)\mathbf{R}_{j, t_2}^{\top}&=\left(\sum_{i=1}^{t_1}\sum_{j=1}^{t_2-1}\mathbf{R}_{i, t_1}\mbox{Cov}(\pmb{e}_i,\pmb{e}_j)\mathbf{R}_{j, t_2-1}^{\top}\right)\mathbf{R}_{t_2, t_2}^{\top}\nonumber\\
&\quad+\left(\sum_{i=1}^{t_1}\mathbf{R}_{i, t_1}\mbox{Cov}(\pmb{e}_i,\pmb{e}_{t_2})\right)\mathbf{R}_{t_2, t_2}^{\top}.
\end{align*}
We can see that $\sum_{i=1}^{t_1}\sum_{j=1}^{t_2-1}\mathbf{R}_{i, t_1}\mbox{Cov}(\pmb{e}_i,\pmb{e}_j)\mathbf{R}_{j, t_2-1}^{\top}$ is the first part of the $\mbox{Cov}(\pmb{s}_{t_1+1}, \pmb{s}_{t_2})$ and the $\sum_{i=1}^{t_1}\mathbf{R}_{i, t_1}\mbox{Cov}(\pmb{e}_i,\pmb{e}_{t_2})$ is the second part of $\mbox{Cov}(\pmb{s}_{t_1+1}, \pmb{s}_{t_2})$. For the second part of the $\mbox{Cov}(\pmb{s}_{t_1+1}, \pmb{s}_{t_2+1})$, 

\begin{equation*}
\begin{aligned}
\sum_{i=1}^{t_1}\mathbf{R}_{i, t_1}\mbox{Cov}(\pmb{e}_i,\pmb{e}_{t_2+1})&=\mathbf{R}_{t_1, t_1}\left(\sum_{i=1}^{t_1-1}\mathbf{R}_{i, t_1-1}\mbox{Cov}(\pmb{e}_i,\pmb{e}_{t_2+1})\right)\\
&\quad+\mathbf{R}_{t_1, t_1}\mbox{Cov}(\pmb{e}_{t_1},\pmb{e}_{t_2+1}).
\end{aligned}
\end{equation*}
 We can find that $\sum_{i=1}^{t_1-1}\mathbf{R}_{i, t_1-1}\mbox{Cov}(\pmb{e}_i,\pmb{e}_{t_2+1})$ is the second part of the $\mbox{Cov}(\pmb{s}_{t_1}, \pmb{s}_{t_2+1})$. For the third part of the $\mbox{Cov}(\pmb{s}_{t_1+1}, \pmb{s}_{t_2+1})$, we have 
\begin{equation*}
\begin{aligned}
\sum_{i=1}^{t_2}\mbox{Cov}(\pmb{e}_{t_1+1},\pmb{e}_i)\mathbf{R}_{i, t_2}^{\top}=\left(\sum_{i=1}^{t_2-1}\mbox{Cov}(\pmb{e}_{t_1+1},\pmb{e}_i)\mathbf{R}_{i, t_2-1}^{
\top}\right)\mathbf{R}_{t_2, t_2}^{\top}
+\mbox{Cov}(\pmb{e}_{t_1+1},\pmb{e}_{t_2})\mathbf{R}_{t_2, t_2}^{\top}.
\end{aligned}
\end{equation*}
Similarly, We can see that $\sum_{i=1}^{t_2-1}\mathbf{R}_{i, t_2}\mbox{Cov}(\pmb{e}_i,\pmb{e}_{t_1+1})$ is the third part of the $\mbox{Cov}(\pmb{s}_{t_1+1}, \pmb{s}_{t_2})$.
Therefore, we can reuse the intermediate results in calculating $\mbox{Cov}(\pmb{s}_{t_1+1}, \pmb{s}_{t_2})$ and $\mbox{Cov}(\pmb{s}_{t_1}, \pmb{s}_{t_2+1})$ to calculate $\mbox{Cov}(\pmb{s}_{t_1+1}, \pmb{s}_{t_2+1})$. The detailed algorithm is summarized in Algorithm~\ref{algovarsv1}.

\begin{algorithm}[!t]
\linespread{1.2}\selectfont
Input: the policy parameters $\pmb{\theta}=\{\pmb{\theta}_1, \pmb{\theta}_2,...,\pmb{\theta}_{H-1}\}$, the covariance matrix between $\pmb{e}_i$ and $\pmb{e}_j$, the parameters of pKG model $\left\{\pmb{\beta}^{s}_{1:H-1}, \pmb{\beta}^{a}_{1:H-1}\right\}$, the rewards coefficients $\pmb{b}_{1:H}$ and $\pmb{c}_{1:H}$\\
(1) Calculate all the covariance matrices:\\
$\mathbf{R}_{0,0}=\pmb{0}$;
$\mathbf{A}^i_{0, t} = \mathbf{A}^i_{t, 0} = \mathbf{0}$ for $t = 1, 2,...,H$ and $i =1,2,3$;\\
\For{$t_2=0, 1,\ldots, H-1$}{
\For{$t_1=0, 1, \ldots, t_2$}{$
\mathbf{A}^{1\top}_{t_2+1, t_1+1}=\mathbf{A}^1_{t_1+1, t_2+1} = (\mathbf{A}^1_{t_1+1, t_2}+\mathbf{A}^2_{t_1+1, t_2})\mathbf{R}_{t_2, t_2}^{\top}$;\
$\mathbf{A}^{2\top}_{t_2+1, t_1+1}=\mathbf{A}^2_{t_1+1, t_2+1} = \mathbf{R}_{t_1, t_1}(\mathbf{A}^2_{t_1, t_2+1}+\mbox{Cov}(\pmb{e}_{t_1}, \pmb{e}_{t_2+1}))$;\
$\mathbf{A}^{3\top}_{t_2+1, t_1+1} = \mathbf{A}^3_{t_1+1, t_2+1} = (\mathbf{A}^3_{t_1+1, t_2}+\mbox{Cov}( \pmb{e}_{t_1+1},\pmb{e}_{t_2}))\mathbf{R}_{t_2, t_2}^{\top}$;\
$\mbox{Cov}^{\top}(\pmb{s}_{t_2+1}, \pmb{s}_{t_1+1})=\mbox{Cov}(\pmb{s}_{t_1+1}, \pmb{s}_{t_2+1}) =\mathbf{A}^1_{t_1+1, t_2+1}$\\
$+\mathbf{A}^2_{t_1+1, t_2+1}+\mathbf{A}^3_{t_1+1, t_2+1}+\mbox{Cov}(\pmb{e}_{t_1+1}, \pmb{e}_{t_2+1})$.
}}

(2) Calculate the variance of the cumulative reward:\\
$\mbox{Var}(\sum_{t=0}^{H-1} r_{t+1})=0$;\\
\For{$t_2=0, 1,\ldots, H-1$}{
\For{$t_1=0, 1,\ldots, H-1$}{
$\mbox{Var}(\sum_{t=0}^{H-1} r_{t+1}) =\mbox{Var}(\sum_{t=0}^{H-1} r_{t+1})+ \pmb{\alpha}_{t_1+1}\mbox{Cov}(\pmb{s}_{t_1+1},\pmb{s}_{t_2+1})\pmb{\alpha}_{t_2+1}^{\top}$;
}
}
\caption{Variance Calculation for Given Policy Parameters}\label{algovarsv1}
\end{algorithm}

\begin{proposition}
\label{prop: time complexity analysis for variance-based SV}
Given the sample size of the model parameters $Q$ and permutations $D$, the cost to compute the approximate variance-based SV of policy parameters with value function in Equation~(\ref{eq: value_theta_var}) for linear Gaussian pKG using Algorithm~\ref{algovarsv1} is $O\left(QDH^3(n^4m+n^3m^2)\right)$.  For the brute force algorithm without reusing calculations, the cost is $O\left(QDH^6(n^4m+n^3m^2)\right)$.
\end{proposition}

Proposition~\ref{prop: time complexity analysis for variance-based SV} shows the time complexity for variance-based Shapley analysis with the reusing of calculation, where the overall procedure is summarized as Algorithm~\ref{agm: sv_linear_variance} in Appendix~\ref{sec: agm_policy}. Compared to the brute force algorithm that calculates the value function directly, the algorithm with reusing of calculation can save the computational cost by a factor of $O(H^3)$. The detailed proof can be found in Appendix~\ref{subsecA: proof_prop6}. When the time horizon $H$ is large, which is common in online biomanufacturing, our algorithm can increase the computational efficiency significantly.

\subsection{SV Estimation for Model Parameters}
\label{sec: linear_modelsv}
Now we focus on the SV estimation for the model parameters, i.e., $\mathcal{O} = \{w^i_t, i=1,\ldots,\mathcal{I}, t=1,\ldots,H\}$. Let $Y=\sum_{t=0}^{H-1} r_{t+1}$, $\pmb{w}_{\mathcal{O}/\mathcal{U}}$ denote the model parameters that are not in subset $\mathcal{U}$, and $\pmb{w}_{\mathcal{U}}$ denote the remaining model parameters. For the predictive analysis, the value function defined in Equation~(\ref{eq: value_model_predictive}) now becomes 
\begin{equation}
\label{eq: linear_value_model_predictive}
g(\mathcal{U}\vert\pmb{w})=\mathbb{E}[Y\vert\mathcal{U}]=\mathbb{E}\left\{\mathbb{E}[Y\vert\pmb{w}_{\mathcal{U}}, \pmb{w}_{\mathcal{O}/\mathcal{U}}]\right\},
\end{equation}
where the outer expectation is taken on the randomness in $\pmb{w}_{\mathcal{O}/\mathcal{U}}$. The inner expectation $\mathbb{E}[Y\vert\pmb{w}_{\mathcal{U}}, \pmb{w}_{\mathcal{O}/\mathcal{U}}]$ can be written in the same way as Equation~(\ref{eq: linear_value_policy_predictive}), except that the condition $\pmb{\theta}_{\mathcal{O}/\mathcal{U}}=0$ is omitted. Therefore, the calculation reusing in Equation~(\ref{eq: nested_propogation}) can also be applied to reduce the computational cost in the predictive analysis of model parameters.
For the variance-based analysis, the value function $g(\mathcal{U})=\mathbb{E}\left[\mbox{Var}\left[Y\vert\mathcal{O}/\mathcal{U}\right]\right]$.   Then the inner level variance function can be decomposed as
\begin{equation}
\label{eq: var_decompose}
    \mbox{Var}\left[Y\vert\mathcal{O}/\mathcal{U}\right] = \mbox{E}\left\{\mbox{Var}\left[Y\vert\pmb{w}_{\mathcal{U}},\pmb{w}_{\mathcal{O}/\mathcal{U}}\right]\right\} + \mbox{Var}\left\{\mbox{E}\left[Y\vert\pmb{w}_{\mathcal{U}},\pmb{w}_{\mathcal{O}/\mathcal{U}}\right]\right\}.
\end{equation}
In Equation~(\ref{eq: var_decompose}), $\mbox{Var}\left[\sum_{t=0}^{H-1} r_{t+1}\vert\pmb{w}_{\mathcal{U}},\pmb{w}_{\mathcal{O}/\mathcal{U}}\right]$ can be derived analytically as in Equation~(\ref{eq: linear_value_theta_var}) and~(\ref{eq: linear_policy_cov}), and $\mbox{E}\left[\sum_{t=0}^{H-1} r_{t+1}\vert\pmb{w}_{\mathcal{U}},\pmb{w}_{\mathcal{O}/\mathcal{U}}\right]$ can be calculated as  Equation~(\ref{eq: linear_value_policy_predictive}). Therefore, the calculation reusing method in Algorithm~\ref{algovarsv1} and Equation~(\ref{eq: nested_propogation}) for the variance-based analysis of policy parameters can also be applied to the model parameters to save the computational cost.


\section{Permutation Sampling Method}
\label{sec: sampling}


In this section, we provide a procedure for sampling well-distributed high-dimensional permutations (called TFWW-VRT), which can be used for SV estimation in both general nonlinear and linear Gaussian pKG models. 
As the name suggests, TFWW-VRT enhances the permutation sampling method in \cite{mitchell2022sampling} in the following two aspects. Firstly, it introduces an alternative hypersphere sampling method, called TFWW transformation in \cite{fang1993number}, to sample the permutations, which has not been used for permutation sampling. We demonstrate that TFWW can be employed to generate permutations more uniformly than BMT and more efficiently than SCT.
Secondly, it integrates the hypersphere sampling with two variance reduction techniques: randomized QMC and AS, which can help produce permutations that are more representative of the whole permutation space and significantly improve the estimation accuracy of the SV.

Algorithm~\ref{algosampling} illustrates TFWW-VRT for generating $D$ samples of $s$-dimensional permutations, i.e., lines 6-7 in Algorithm~\ref{agm: svgeneral}, which consists of four steps. In step 1, assuming that $D=2d$, we generate $d$ points from $[0,1]^{s-2}$ using randomized QMC; Step 2 transforms these $d$
points to a $(s-2)$-sphere using TFWW transformation; In step 3, we map the spherical points to a $s$-dimensional permutation space using the method in \cite{mitchell2022sampling}; In step 4, we generate the antithetic samples of $d$ permutations to obtain total $D$ permutation samples.
\SetInd{0.5em}{1em}
\SetNlSkip{1em}
\begin{algorithm}[t!]
\linespread{1.2}\selectfont
Input: the number of permutations $D=2d$, the dimension of the permutation $s$\

Step 1. Generate $d$ points from $[0,1]^{s-2}$ using randomized Sobol’ sequence\;
Step 2. Transform the points to the $(s-2)$-sphere using TFWW transformation \\
\quad procedure in Appendix~\ref{subsec: Appendix_tfww} and obtain $\{\pmb{x}^{(i)}, i = 1,2,\ldots,d\}$;\\
Step 3. Map the points to the permutations:\\
\For{$i=1,2,\ldots,d$}{
$\pmb{x}'^{(i)} = \hat{T}^\top\pmb{x}^{(i)}$, where $\pmb{x}^{(i)}$ is the $i$th point obtained in step 2,\\
$\hat{T}=\left[\begin{array}{ccccc}
\frac{1}{\sqrt{2}} & -\frac{1}{\sqrt{2}} & 0 & \ldots & 0 \\
\frac{1}{\sqrt{6}} & \frac{1}{\sqrt{6}} & -\frac{2}{\sqrt{6}} & \ldots & 0 \\
& \vdots & & \ddots & \\
\frac{1}{\sqrt{s^2-s}} & \frac{1}{\sqrt{s^2-s}} & \frac{1}{\sqrt{s^2-s}} & \ldots & -\frac{s-1}{\sqrt{s^2-s}}
\end{array}\right]_{(s-1)\times s}$\\
$\pi^{(i)} = \mbox{argsort}(\pmb{x}'^{(i)})$, where $\mbox{argsort}(\pmb{x}')=\pmb{b}$ means $x'_{b_1} \le x'_{b_2} \ldots \le x'_{b_s}$;
}

Step 4. Generate the antithetic permutations using Equation~(\ref{eq: reversion}):\\
\quad $\bar{\pi}^{(i)} = \mbox{Reverse}(\pi^{(i)})$ for $i = 1,2,...,d$;\\
Return $D$ permutations $\{\pi^{(i)}, \bar{\pi}^{(i)}\}_{i=1}^d$.
\caption{TFWW-VRT}\label{algosampling}
\end{algorithm}

We first focus on sampling the points uniformly from a unit hypersphere with TFWW tranformation. 
Denote $\left(Y_{1}, \ldots, Y_{s}\right) \sim D_{s}\left(\alpha_{1}, \ldots, \alpha_{s}\right)$ if  $\left(Y_{1}, \ldots, Y_{s}\right)$  has a Dirichlet distribution with parameters  $(\alpha_{1}, \ldots, \alpha_{s})$ and suppose that  $\pmb{x} \sim \mathrm{Unif}\left(U_{s}\right)$. We can sample $\pmb{x}$ based on the following fact \citep{fang1993number}:
\begin{equation}
\begin{aligned}
\pmb{x} \stackrel{d}{=}\left(\begin{array}{c}
d_{1} \pmb{x}_1 \\
\vdots \\
d_{a} \pmb{x}_a
\end{array}\right),
\end{aligned}
\label{eq: TFWW_fact}
\end{equation}
where
\begin{itemize}    
\item[(1)]  $\pmb{x}_j \sim \mathrm{Unif}\left(U_{t_{j}}\right) \mbox{ for } j=1, \ldots, a \mbox{ and } t_{1}+\cdots+t_{a}=s$;
\item[(2)] $d_{j}>0 \mbox{ for } j=1, \ldots, a, \text{ and } \left(d_{1}^{2}, \ldots, d_{a}^{2}\right) \sim D_{a}\left(t_{1} / 2, \ldots\right. ,  \left.t_{a} / 2\right)$; 
\item[(3)]  $\left(d_{1}, \cdots, d_{a}\right), \pmb{x}_1, \ldots, \pmb{x}_a$  are independent.
\end{itemize}

Therefore, we can sample  $\pmb{x}$ by generating the samples of $\left(d_{1}, \ldots, d_{a}\right)$ and $ \pmb{x}_1, \ldots, \pmb{x}_a$. Note that $\left(d_{1}, \ldots, d_{a}\right)$ can be generated from the uniform distribution in $C^{a-1}=[0,1]^{a-1}$, denoted by  $\mathrm{Unif}\left(C^{a-1}\right)$ \citep{gelman1995bayesian}. When $s$ is even and $a = s/2$, we  set $t_1 = t_2 = \cdots = t_a = 2$,  then $\{\pmb{x}_1, \ldots, \pmb{x}_a\}$ can be generated from $\mathrm{Unif}(C^{a})$. When $s$ is odd and $a = (s-1)/2$, we  set
$t_1 = 3$, $t_2 = \cdots = t_a = 2$, then $\{\pmb{x}_1, \ldots, \pmb{x}_a\}$ can be generated from $\mathrm{Unif}(C^{a+1})$. 
The detailed TFWW transformation procedure is described in Appendix~\ref{subsec: Appendix_tfww}. 

Note that both TFWW transformation and SCT sampling from a hypersphere based on $\mathrm{Unif}(C^{s-1})$, whereas BMT samples from a hypersphere based on $\mathrm{Unif}(C^{s})$ or $\mathrm{Unif}(C^{s+1})$, which can lead to higher discrepancy of the final permutations.

In addition, 
the time complexity for different sampling methods is summarized in Proposition~\ref{prop: time complexity analysis for permutation sampling}. 
This proposition shows that the computational costs of the BMT method and TFWW method grow linearly with the dimension of the hyperspere $s$, and the computational cost of SCT grows quadratically with~$s$. The proof of  Proposition~\ref{prop: time complexity analysis for permutation sampling} is shown in Appendix~\ref{subsecA: proof_prop4}. Therefore, TFWW transformation can balance both the discrepancy of the final permutations and the running time of the algorithm, which is further validated numerically in Section~\ref{subsec: performance}.

\begin{proposition}
\label{prop: time complexity analysis for permutation sampling}
 If we denote the number of permutations as $D$, the dimension of the permutations as $s$, the time complexities for the three methods are shown as follows: 
{
\linespread{1.0} \selectfont
\begin{equation}
\begin{array}{ll}
\hline \text {Algorithm\quad\quad} & \text {Complexity} \\
\hline \text {BMT} & O\left(Ds\right) \\
\text {SCT} & O(Ds^2) \\
\text {TFWW} & O\left(Ds\right) \\
\hline
\end{array}
\nonumber
\end{equation}}
\end{proposition}

A limiting factor in the accuracy of a standard random or pseudo-random sequence used to generate points from $\mathrm{Unif}\left(U_{s}\right)$ is the clumping and the empty spaces that occur among the points. To generate more uniformly distributed permutations, we use the randomized
Sobol' sequence, a low-discrepancy sequence, to generate the points in the unit hypercube \citep{l2018randomized}. Note that the randomized Sobol’ sequence is widely used in QMC and is relatively inexpensive in practice \citep{leovey2015quasi}. 

To further improve the estimation, we combine the AS, which is a variance reduction technique for Monte Carlo integration where samples are taken as correlated pairs instead of standard i.i.d. samples \citep{Lemieux2009}. Note that AS for permutations can be a simple strategy that involves taking permutations and their reverse, i.e., the antithetic Monte Carlo estimator is
\begin{equation}
\begin{aligned}
\hat{\mu}_{\text {anti }}=\frac{1}{D} \sum_{i=1}^{D / 2} (f\left(\pi^{(i)}\right)+f\left(\mbox{Reverse}(\pi^{(i)})\right)),
\nonumber
\end{aligned}
\end{equation}
where 
\begin{equation}
\begin{aligned}
\mbox{Reverse}(\pi^{(i)}) = \{\pi^{(i)}_s,\pi^{(i)}_{s-1},\ldots,\pi^{(i)}_1\} \text{ if } \pi^{(i)} = \{\pi^{(i)}_1,\pi^{(i)}_2,\ldots,\pi^{(i)}_s\}.
\end{aligned}
\label{eq: reversion}
\end{equation}
 We find that such method can further improve our estimation accuracy, as used in Step 4 in Algorithm~\ref{algosampling}.



\section{Empirical Study}
\label{sec:empiricalStudy}


\begin{figure}[!b]
\centering
\includegraphics[width=1\textwidth]{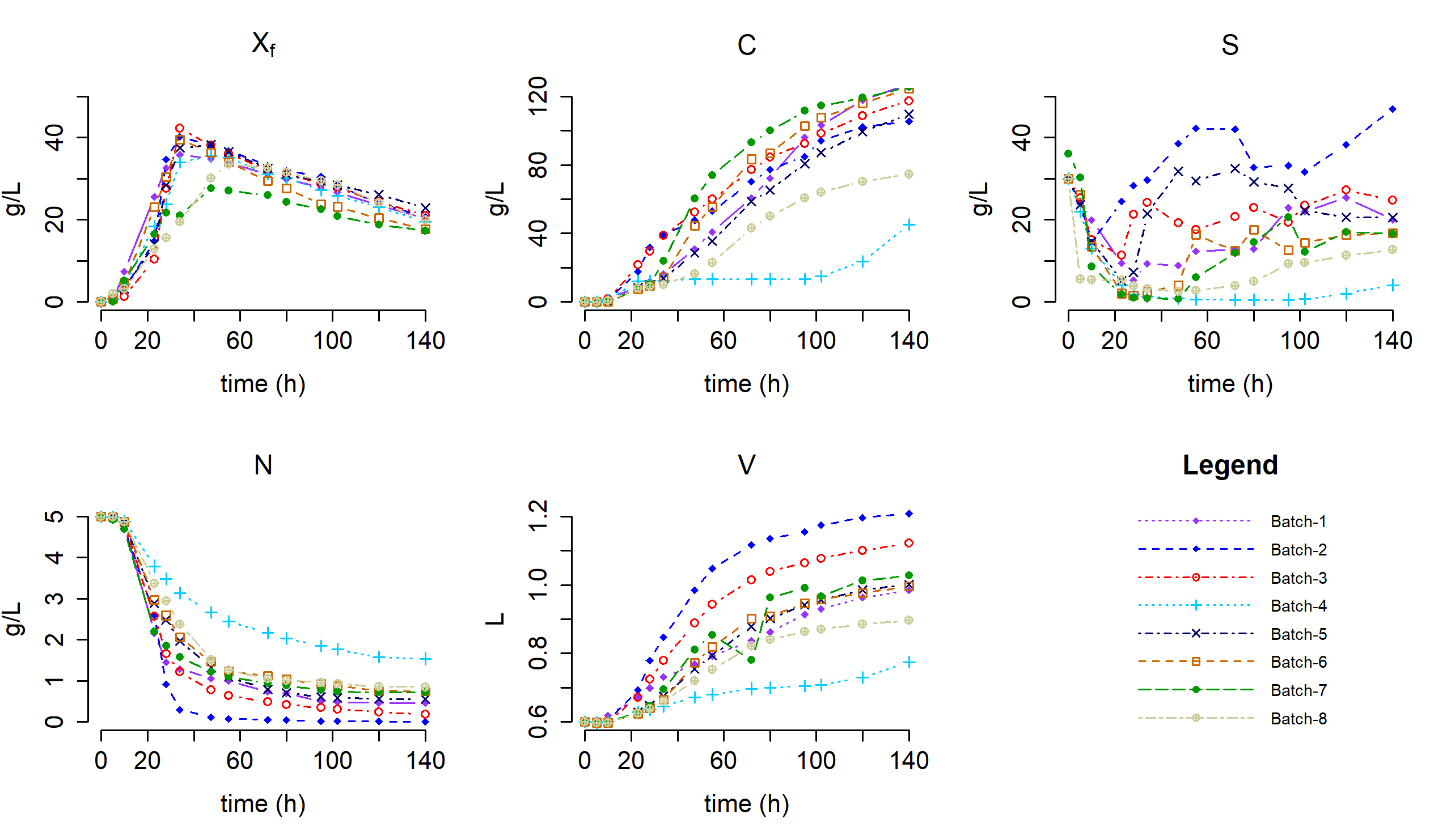}
\caption{Illustration of 8 real lab experiment trajectory batches in the fermentation process.}\label{fig: real data}
\end{figure}

In this section, 
we present a case application of our approach to multiphase
fermentation of Yarrowia lipolytica as briefly mentioned in Example~\ref{example-1}, a process in
which viable cells grow and produce the biological
substance of interest. We first describe the data and the bioprocess in Section~\ref{subsec: data}. Then in Section~\ref{subsec: performance}, we show the effectiveness and efficiency of TFWW-VRT permutation sampling algorithm compared with the BMT and SCT algorithms. Finally, we conduct SV analysis for the general nonlinear pKG models in Section~\ref{subsec: nonlinear sv analysis} and for the linear Gaussian pKG model in Section~\ref{subsec: linear sv analysis}, respectively. We show the computational efficiency of our algorithms and demonstrate how the results can benefit performance improvement and stability control in biomanufacturing. All source code and data are available at \url{https://github.com/zhaojksim/SV_pKG}. 

\subsection{Data and Model Description}
\label{subsec: data}
Our data are from the fed-batch
fermentation process of Yarrowia lipolytica yeast \citep{zheng2022policy}. 
%
As we introduce in Example~\ref{example-1}, the five-dimensional continuous state variable $\pmb{s} = (X_f,C,S,N,V)$ is defined as follows: $X_f$ represents the lipid-free cell mass, $C$ measures citrate, the actual ``product'' to be harvested at the end of the bioprocess, generated by the cells' metabolism, $S$ and $N$ are the quantities of substrate (a type of oil) and nitrogen, respectively, used for cell growth and production, and $V$ is the working volume of the entire batch. Additionally, we have a single scalar continuous CPP action $a=F_S$, which denotes the feed rate, or the quantity of new substrate that is supplied to the working cells in a single unit of time.
Figure~\ref{fig: real data} illustrates 8 batches of real lab experimental trajectories. 
We can see that there is a great deal of variation between real experiments.

The kinetic equations from the domain knowledge that describe the dynamics of Yarrowia lipolytica fermentation process are provided in Appendix~\ref{secA2}, in which some model parameters are given in Table~\ref{tab:model parameters}. 
In our experiments, we consider both general nonlinear pKG model and linear Gaussian pKG model as described in Sections~\ref{sec: general model} and~\ref{sec: linear Gaussian}, respectively.

For the general nonlinear pKG model, we transform the kinetic equations in Appendix~\ref{secA2} to the nonlinear transition model using the method described in Example~1. Following the real data in Figure~\ref{fig: real data}, the time horizon $H = 8$. The model parameters in the general nonlinear pKG model are $\pmb{w} = \{m_s, r_L, \beta_{LCmax}, \mu_{max}\}$, whose prior distribution is set to be Gaussian and the posterior distribution can be learned from the data. Other model parameters are fixed as in Table~\ref{tab:model parameters}. To ensure feasible action, we adopt the policy as $\pmb{a}_t  = \max\{\pmb\mu^a_t + \pmb{\theta}_t^\top(\pmb{s}_t - \pmb\mu^{s}_t), 0\}$,
which is a simple nonlinear policy and the parameters $\pmb\mu^a_t$ and $\pmb\mu^{s}_t$ are set by ourselves. The reward function is set to be 
\begin{equation*}
  r(\pmb{s}_t, a_t)=\begin{cases}
    -15+1.29C_t, & \text{if $t = H$}.\\
    -534.52a_t, & \text{otherwise},
  \end{cases}
\end{equation*}
which means that, during the process,
the operating cost is primarily driven by the substrate,
and the product revenue is collected at the harvest
time.

For the linear Gaussian pKG model, we use the linear Gaussian structure of pKG model described in Section~\ref{sec: linear Gaussian}. The time horizon $H$ is set to be 36 and there are 5 CQA state nodes and one CPP action node per time period, which leads to 175 policy parameters. To mimic “real-world data” collection, we generate
the data according to \cite{zheng2022policy}.

\subsection{Performance of Permutation Sampling Methods}
\label{subsec: performance}
   
In this section, we first show the superiority of the TFWW hypersphere transformation over the SCT and BMT methods based on the uniformity of the generated permutation samples and the running time. Then we evaluate the performance of different sampling methods in terms of SV estimation error.

\subsubsection{Uniformity and Efficiency of TFWW Transformation.}

\begin{table}[!t]
\centering
\caption{Discrepency score of sampled permutations using different permutation sampling methods.}\label{tab: discrepancy}
\vspace{0.3cm}
\begin{tabular}{clrr}
    \hline
\multicolumn{1}{c}{Dimension}&
    \multicolumn{1}{c}{Method} & \multicolumn{1}{c}{Mean} & \multicolumn{1}{c}{Std} \\
    \hline
     \multirow{3}[0]{*}{5} & BMT & 0.0571  & 0.0052  \\
          & SCT   & 0.0517  & 0.0031  \\
          & TFWW  & 0.0476  & 0.0034  \\
    \cline{1-4}
    \multirow{3}[0]{*}{10} & BMT & 0.0784  & 0.0018  \\
          & SCT   & 0.0733  & 0.0012  \\
          & TFWW  & 0.0727  & 0.0013  \\
    \cline{1-4}
    \multirow{3}[0]{*}{20} & BMT & 0.0841  & 0.0015  \\
          & SCT   & 0.0809  & 0.0005  \\
          & TFWW  & 0.0808  & 0.0006  \\
    \cline{1-4}
    \multirow{3}[0]{*}{30} & BMT & 0.0858  & 0.0007  \\
          & SCT   & 0.0830  & 0.0005  \\
          & TFWW  & 0.0830  & 0.0004  \\
    \cline{1-4}
    \multirow{3}[0]{*}{40} & BMT & 0.0868  & 0.0006  \\
          & SCT   & 0.0838  & 0.0003  \\
          & TFWW  & 0.0838  & 0.0003  \\
    \hline
    \end{tabular}%
\end{table}%

To measure the uniformity of the generated permutations, we use the discrepancy score 
suggested by \cite{mitchell2022sampling}. We fix the number of samples as 100 and run the permutation sampling with three permutation methods for different dimensions and calculate the mean and the standard deviation of the discrepancy score over 36 macro-replications. The results for different dimensions are shown in Table~\ref{tab: discrepancy}. Note that the higher the discrepancy score, the worse the uniformity. As we can see, BMT method has a higher discrepancy than the other two methods in all dimensions. As the inputs of the BMT are random points of higher dimension than the other two methods, it would result in more clusters or empty spaces. The TFWW method and the SCT method show similar performance in all dimensions except in the case when the dimension of the permutation $s = 5$. 

\begin{table}[!t]
\linespread{1}\selectfont
  \centering
  \caption{Running time (in seconds) of different permutation sampling methods.}\label{tab: time}
  \vspace{0.3cm}
  \begin{tabular}{cclrr}
    \hline
    Dimension & \# of samples & \multicolumn{1}{c}{Method} & \multicolumn{1}{c}{Sobol generation} & \multicolumn{1}{c}{Transformation} \\
    \hline
    \multirow{9}[0]{*}{20} & \multirow{3}[0]{*}{1000} & BMT & 0.284  & 0.001  \\
          &       & SCT   & 0.258 & 3.200 \\
          &       & TFWW  & 0.264  & 0.001  \\
          \cline{2-5}
          & \multirow{3}[0]{*}{2000} & BMT & 0.292  & 0.001  \\
          &       & SCT   & 0.255 & 6.465 \\
          &       & TFWW  & 0.257  & 0.001  \\
          \cline{2-5}
          & \multirow{3}[0]{*}{4000} & BMT & 0.291  & 0.003  \\
          &       & SCT   & 0.256 & 13.130 \\
          &       & TFWW  & 0.257  & 0.002  \\
          \cline{1-5}
    \multirow{9}[0]{*}{40} & \multirow{3}[0]{*}{1000} & BMT & 0.582  & 0.002  \\
          &       & SCT   & 0.549 & 13.932 \\
          &       & TFWW  & 0.562  & 0.002  \\
          \cline{2-5}
          & \multirow{3}[0]{*}{2000} & BMT & 0.573  & 0.003  \\
          &       & SCT   & 0.552 & 30.063 \\
          &       & TFWW  & 0.543  & 0.003  \\
          \cline{2-5}
          & \multirow{3}[0]{*}{4000} & BMT & 0.622  & 0.005  \\
          &       & SCT   & 0.610 & 59.553 \\
          &       & TFWW  & 0.606  & 0.006  \\
    \cline{1-5}
    \multirow{9}[0]{*}{80} & \multirow{3}[0]{*}{1000} & BMT & 1.234  & 0.004  \\
          &       & SCT   & 1.153 & 62.510 \\
          &       & TFWW  & 1.164  & 0.003  \\
          \cline{2-5}
          & \multirow{3}[0]{*}{2000} & BMT & 1.211  & 0.006  \\
          &       & SCT   & 1.148 & 123.862 \\
          &       & TFWW  & 1.131  & 0.005  \\
          \cline{2-5}
          & \multirow{3}[0]{*}{4000} & BMT & 1.263  & 0.012  \\
          &       & SCT   & 1.179 & 250.936 \\
          &       & TFWW  & 1.179  & 0.010  \\
          \hline
    \end{tabular}
\end{table}%

We then record the running time for three methods and calculate the average time over 36~macro-replications in Table~\ref{tab: time}.
The average time of randomized Sobol' sequence generation is listed in column 4 and the average running time of each transformation method (i.e., BMT, SCT, TFWW) is listed in column 5. First, we find that the SCT method runs significantly slower compared with the other two methods. The running time of the SCT method increases linearly with the number of samples but quadratically with the dimension of permutation. For example, when we fix the dimension of the permutation as 20 and change the number of samples from 1000 to 2000, the running time of the SCT method is doubled (from 3.200s to 6.465s). When we fix the number of samples as 1000 and change the dimension from 20 to 40, the running time of the SCT method is quadrupled (from 3.200s to 13.932s). Therefore,
the SCT method is expensive in higher dimensions. Second, the running times of the BMT and TFWW transformation grow linearly with both the number of samples and the dimension of permutation, which are negligible compared with the Sobol' sequence generation procedure and much faster than the SCT method. Combining the results above, TFWW shows better performance in terms of both permutation uniformity and running time. 

\subsubsection{Estimation Error of TFWW-VRT.}

We also evaluate the performance of different permutation sampling methods in the SV estimation of pKG models. The model here we test on is the linear Gaussian pKG model as described in Section~\ref{subsec: data}. We conduct predictive analysis for policy parameters using four permutation sampling methods: the BMT and SCT methods from \mbox{\cite{mitchell2022sampling}}, the TFWW method with and without randomized QMC and AS (TFWW and TFWW-VRT, respectively). To measure their estimation performance and balance the computational time, we follow the experimental framework in \mbox{\cite{mitchell2022sampling}} and choose 10 sets of model parameters. For the $q$th set of model parameters, $\pmb{w}^{(q)}$, we sample permutations using 4 different methods and calculate SVs for policy parameters. The resulting estimation error is quantified by the mean squared error (MSE), which is defined as follows:
\begin{equation}
\begin{aligned}
\mbox{MSE}=\frac{1}{1750}\sum_{q=1}^{10}\sum_{i=1}^{175}(\mbox{Sh}(\theta_i\vert \pmb{w}^{(q)})-\hat{\mbox{Sh}}(\theta_i\vert \pmb{w}^{(q)}))^2,
\nonumber
\end{aligned}
\end{equation}
where $\mbox{Sh}(\theta_i\vert\pmb{w}^{(q)})$ is the exact SV of the $i$th dimension of the policy parameters $\theta_i$ given $\pmb{w}^{(q)}$, which is estimated using 20,000 random permutations, and $\hat{\mbox{Sh}}(\theta_i\vert \pmb{w}^{(q)})$ is the estimated SV given $\pmb{w}^{(q)}$. We repeat the experiment 25 times to generate 95\% confidence intervals.

The experimental results are shown in Figure~\ref{fig: mse}. First, we can see TFWW performs better than SCT and BMT in terms of estimation error. Therefore, combined with the results in Table~\ref{tab: time}, TFWW is capable 
 of achieving computational efficiency while maintaining high estimation accuracy. Second, we find that our proposed TFWW-VRT, i.e., TFWW with randomized QMC and AS, can significantly enhance the SV estimation accuracy under the scenario of limited permutation samples, which indicates the significance of variance reduction in SV estimation when the limited computational budget is available.


\begin{figure}[t!]
\centering
\includegraphics[width=0.7\textwidth]{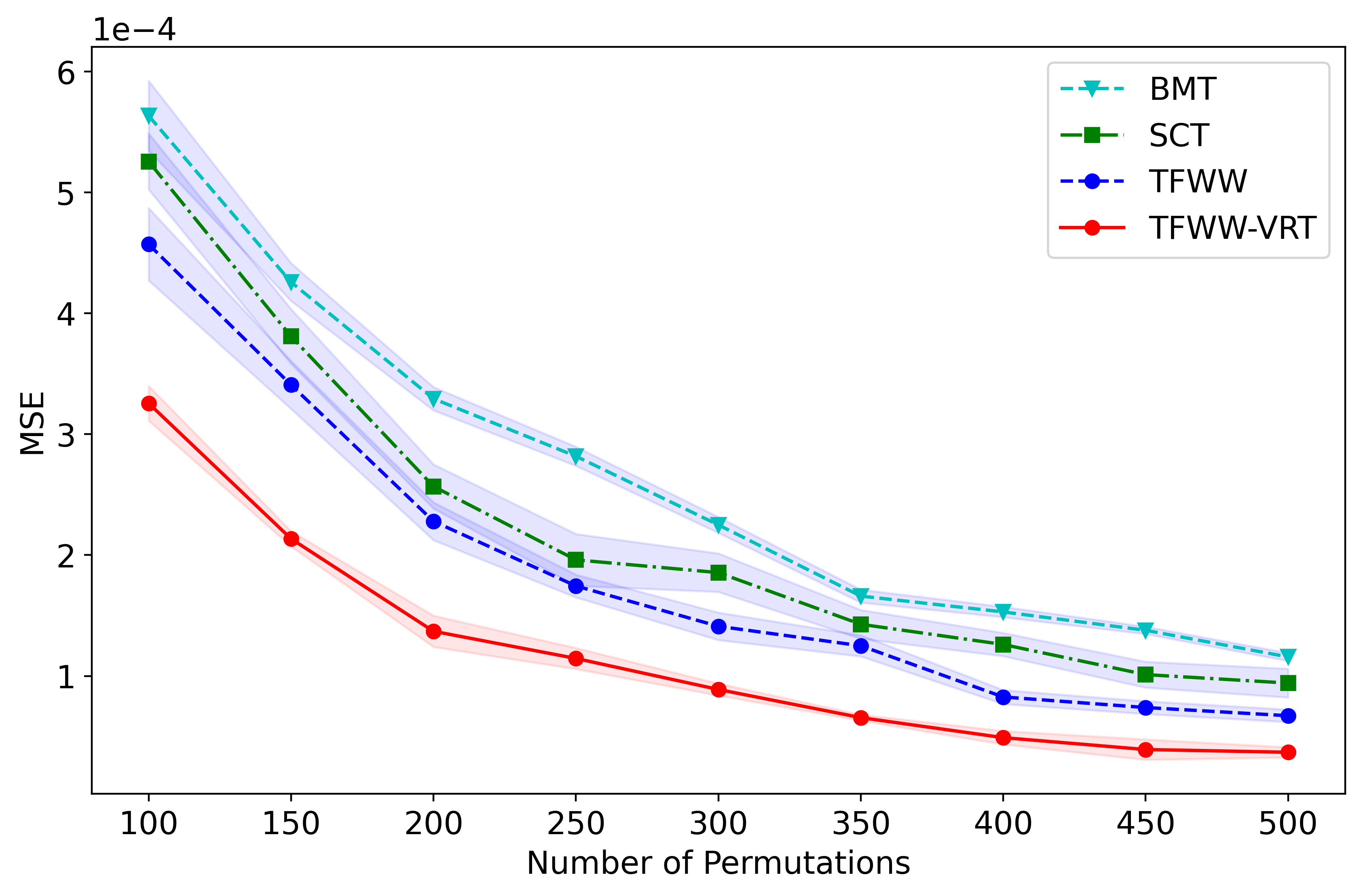}
\caption{Estimation errors for different permutation sampling methods with 95\% confidence interval.}\label{fig: mse}
\end{figure}


\subsection{Empirical Results for General Nonlinear pKG Model}
\label{subsec: nonlinear sv analysis}
In this subsection, we illustrate how the SV can benefit the sensitivity and interpretability analysis of the biomanufacturing process. We want
to study how the inputs at different time horizons contribute to the expectation and the variation
of the cumulative rewards $J$ and final citrate output $C_H$ while considering the model uncertainty. We calculate the SV for the general nonlinear pKG model in Section~\mbox{\ref{subsec: data}} using the procedures described in Section~\ref{sec: nonlinear_SV}. The SVs of the random factors on the expectation of the outputs are calculated by setting $e^k_h = \mathbb{E}\left[e^k_h\right]+\sigma(e^k_h)$, where $\mathbb{E}(e^k_h)$ denotes the expectation of $e^k_h$ and $\sigma(e^k_h)$ denotes the standard deviation of $e^k_h$, which means the random factor is set as a positive deviation from their means. The sample size is determined by Equation~(\ref{eq: sample_size_sigma}), where the variance of the marginal contribution is estimated from the pilot experiment.

We first compare the efficiency of our algorithm, which reuses the calculation of the pathway, and the brute force algorithm, which directly estimates the SV and does not utilize the structure information of pKG model. For the general nonlinear pKG models, the number of state $n = 5$ and the number of action $m = 1$. The number of permutations $D = 1000$ and the sample size of model parameters $Q = 1$ for comparison. We record the running time of the SV estimation with varying time horizons for the policy parameters in Table~\ref{tab: nonlinear efficiency}, which is averaged over 10 macro-replications. The results show that the reuse of calculation can save the computation significantly, which allows large-scale sensitivity analysis in the biomanufacturing process.

\begin{table}[!t]
\begin{center}
  \caption{Computational time (in seconds) of proposed and brute force algorithm for general nonlinear pKG model.}\label{tab: nonlinear efficiency}
  \vspace{0.3cm}
    \begin{tabular}{lrrrr}
    \hline
          \multicolumn{1}{c}{Time Horizon} & \multicolumn{1}{c}{Proposed} & \multicolumn{1}{c}{Brute force}\\
    \hline
    H=4   & 70.42  &  149.58\\
    H=8   & 377.86  &  1561.33\\
    H=16  & 1757.13  &  13535.94\\
\hline
\end{tabular}
\end{center}
\end{table}%

\begin{figure}[!b]
\centering
\includegraphics[width=0.8\textwidth]{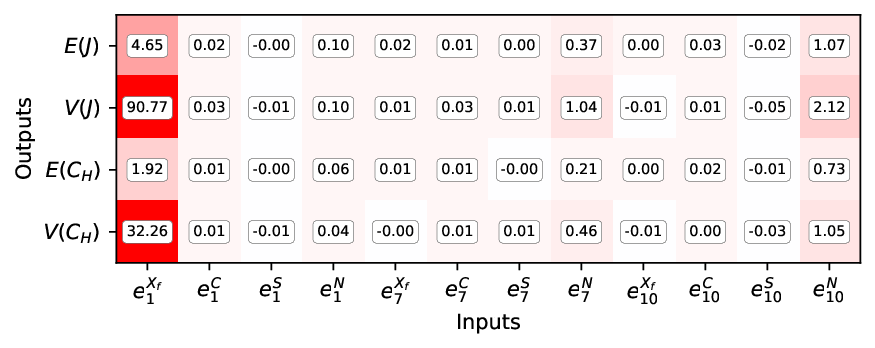}
\caption{SV of selected random factors and outputs in nonlinear pKG model.}\label{fig: nonlinear-random}
\end{figure}

\begin{figure}[t!]
\centering
\includegraphics[width=0.9\textwidth]{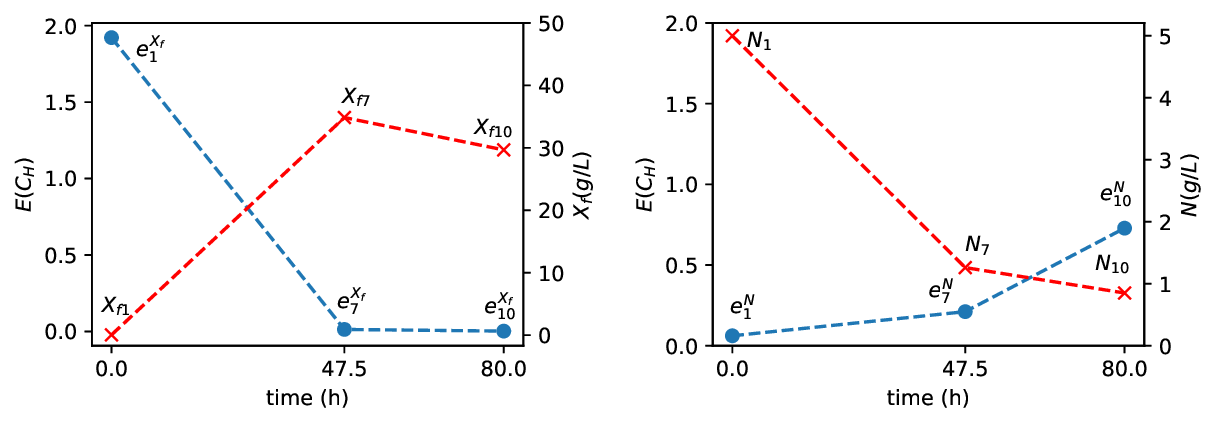}
\caption{SV of random factors and the corresponding state dynamics in real data.}\label{fig: nonlinear sv dynamics}
\end{figure}

Figure~\ref{fig: nonlinear-random} shows the SV of the selected random factors (i.e., $e_t^{x_f}$, $e_t^{C}$, $e_t^{S}$ and $e_t^{N}$ along the horizontal axis) in the general nonlinear pKG model. 
The SVs are colored red. The deeper the red color, the higher the positive value. First, we can see that a positive deviation of the initial cell mass exerts a significant positive impact on the expectation of the cumulative reward~$E(J)$ and the final citrate output $E(C_H)$, which can be seen in column $e^{x_f}_1$. It is consistent with the fact that the initial cell density $X_{f1}$ is relatively low, which can be seen from the average cell mass in real data (the red line at left in Figure~\ref{fig: nonlinear sv dynamics}). Therefore, an increase in cell density can significantly promote the citrate formulation. The initial cell mass also contributes mostly to the variation of the output (i.e., $V(J)$ and $V(C_H)$ in Figure~\ref{fig: nonlinear sv dynamics}), which suggests that controlling the initial cell mass may benefit the improvement of the process stability.
Second, we can also see that the impact of the amount of nitrogen $N_t$ becomes more significant as the fermentation time increases (see the line plot in blue at right in Figure~\ref{fig: nonlinear sv dynamics}). It is because the initial nitrogen $N_1$ is sufficient for the cell growth and citrate formulation. As the bioprocess goes on, the nitrogen is consumed and becomes scarce, which can be observed by the red line at right in Figure~\ref{fig: nonlinear sv dynamics}. Therefore, the amount of nitrogen $N_t$ becomes the bottleneck of the bioprocess and has a greater impact on the cumulative reward $J$ and the final citrate output $C_H$ later in the fermentation process.

\begin{figure}[b!]
\centering
\includegraphics[width=0.8\textwidth]{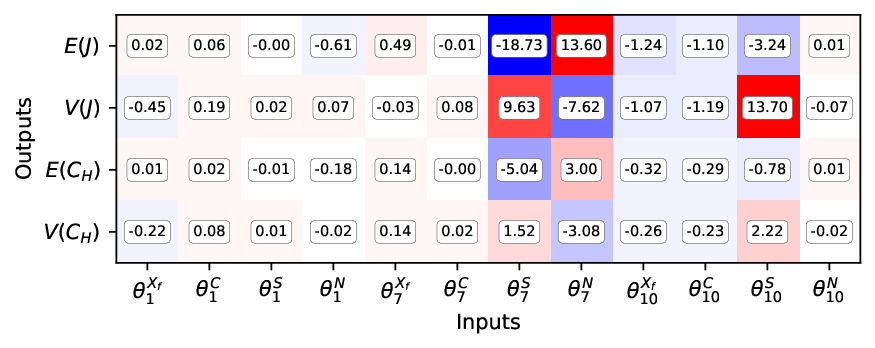}
\caption{SV of selected policy parameters and outputs in the general nonlinear pKG model.}\label{fig: nonlinear-theta}
\end{figure}

Figure~\ref{fig: nonlinear-theta} illustrates the SV of the selected current policy parameters (i.e., $\theta_t^{x_f}$, $\theta_t^{C}$, $\theta_t^{S}$ and $\theta_t^{N}$ along the horizontal axis) in the general nonlinear pKG model. Positive values are in red, while negative values
are in blue. One observation is that the expectation and the variance of the cumulative reward (i.e., $E(J)$ and $V(J)$) and the final citrate output (i.e., $E(C_H)$ and $V(C_H)$) are less affected by the policy parameters at $t = 1$. This observation comes from the sufficient initial substrate concentration, which makes the policy parameters at $t = 1$ less important in the bioprocess. Another observation is that, during the fermentation, the policy parameters related to the substrate contribute significantly to the outputs (see columns $\theta^S_7$ and $\theta^S_{10}$) since it is directly related to the feedback control of the feed rate and the negative contributions arise from the dilution effect of the feed, which can be observed by the dynamic of working volume $V$, i.e., the plot in the middle of the second row of Figure~\ref{fig: real data}. Note that the policy parameters $\theta^{S}_7$ and $\theta^{S}_{10}$ have a negative impact on the expectation of the cumulative reward and the final citrate output and have a positive impact on their variance, the decision-makers need to adjust the current $\theta^{S}_7$ and $\theta^{S}_{10}$ carefully. Finally, we can also see that there are many less important policy parameters in the pKG models, which may provide some insights on the dimension reduction when we try to optimize the pKG models.

\subsection{Empirical Results for Linear Gaussian pKG Model}
\label{subsec: linear sv analysis}

\begin{table}[!t]
\begin{center}
  \caption{Computational time (in seconds) of proposed and brute force algorithm for linear Gaussian pKG model.}\label{tab: linear efficiency}
  \vspace{0.3cm}
    \begin{tabular}{lrrrr}
    \hline
          & \multicolumn{2}{c}{Predictive SV} & \multicolumn{2}{c}{Variance-based SV} \\
          \cmidrule{2-3}\cmidrule{4-5}
          \multicolumn{1}{c}{Time Horizon}
          & \multicolumn{1}{c}{Algorithm~\ref{agm: sv_linear_predictive}} & \multicolumn{1}{c}{Brute force} & \multicolumn{1}{c}{Algorithm~\ref{agm: sv_linear_variance}} & \multicolumn{1}{c}{Brute force} \\
    \hline
    H=4   & 0.24  & 0.25  & 1.10  & 4.54  \\
    H=8   & 1.15  & 3.04  & 11.94  & 332.24  \\
    H=16  & 4.69  & 31.48  & 115.52  & $>15000$ \\
    H=32  & 19.95  & 306.70  & 1000.56  & --------- \\
\hline
\end{tabular}
\end{center}
\end{table}%

We first compare the running time of our algorithm (i.e., Algorithms~\ref{agm: sv_linear_predictive} and~\ref{agm: sv_linear_variance}) and the brute force algorithm in the SV estimation of policy parameters. Table~\ref{tab: linear efficiency} reports the average computational
time in seconds (averaged over 10 macro-replications)
for different time horizons $H$. We can see that our proposed algorithm is much more efficient than the brute force algorithm under the condition that the time horizon is long. When $H$ increases from 8 to 16, the running time of Algorithm~\ref{agm: sv_linear_predictive} for predictive SV increases by about 4 times. However, the running time of the Brute force calculation for predictive SV increases by about 10 times. Therefore, Algorithm~\ref{agm: sv_linear_predictive}} can reduce the cost of computing the estimated SV by a factor of $O(H)$ for predictive SV analysis. For variance-based SV analysis, 
when $H$ increases from 4 to 8, the running time of Algorithm~\ref{agm: sv_linear_variance} for predictive SV increases by about 10 times. However, the running time of the Brute force calculation for variance-based SV increases by about 80 times. Therefore, Algorithm~\ref{agm: sv_linear_variance} can reduce the cost of computing the estimated SV by a factor of $O(H^3)$ for variance-based SV analysis.

\begin{figure}[!b]
\centering
\includegraphics[width=0.8\textwidth]{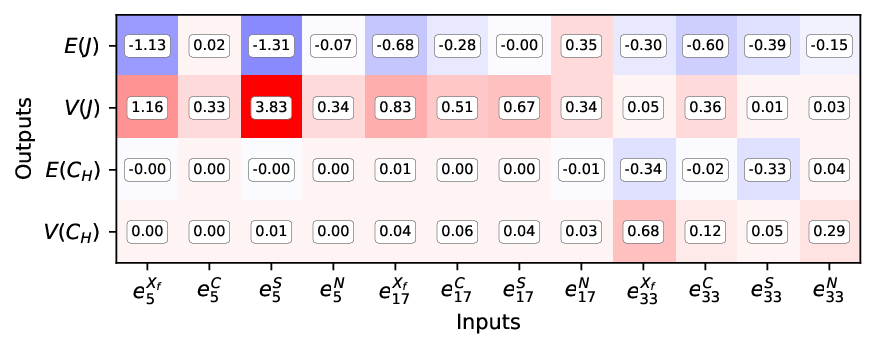}
\caption{SV of selected policy parameters and outputs in linear pKG model.}\label{fig: linear-random}
\end{figure}

Figure~\ref{fig: linear-random} illustrates the SV of the selected random factors (i.e., $e_t^{x_f}$, $e_t^{C}$, $e_t^{S}$ and $e_t^{N}$ along the horizontal axis) in the linear Gaussian pKG model. We can see the random factors $e^{X_f}_5$ and $e^{S}_5$ have a dominant negative impact on the cumulative reward. However, they have less contribution to the final citrate output $C_H$, which suggests that a positive deviation of $e^{X_f}_5$ and $e^{S}_5$ can increase the fermentation cost significantly. The results also show that the random factors $e^{X_f}_{33}$ and $e^{S}_{33}$ have significant negative contributions to the final citrate output $C_H$. Since the citrate formulation is inhibited at high cell density, a positive deviation of $e^{X_f}_{33}$ can decrease the final citrate output. A similar pattern can be seen for $e^{S}_{33}$, whose positive deviation can increase the future cell density and thus decrease the final citrate output. 

\begin{figure}[t!]
\centering
\includegraphics[width=0.8\textwidth]{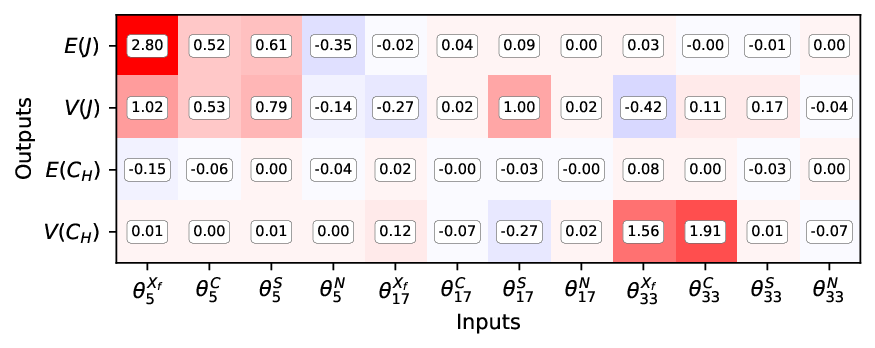}
\caption{SV of selected policy parameters and outputs in linear pKG model.}\label{fig: linear-theta}
\end{figure}

Figure~\ref{fig: linear-theta} shows the SV of the selected current policy parameters (i.e., $\theta_t^{x_f}$, $\theta_t^{C}$, $\theta_t^{S}$ and $\theta_t^{N}$ along the horizontal axis) in the linear Gaussian pKG model. Firstly, we find that the policy parameter $\theta^{X_f}_5$ has a significant positive contribution to the expectation of the cumulative reward, while it has a negative contribution to the final citrate output, which means the  application of the current policy parameter $\theta^{X_f}_5$ can save the fermentation cost. Secondly, we can also see that the policy parameters $\theta^{X_f}_{33}$ and $\theta^{C}_{33}$ have a significant positive impact on the variance of the final citrate output $V(C_H)$ while they contribute less to the expectation of the final citrate output $E(C_H)$, which suggests that the current policy parameters $\theta^{X_f}_{33}$ and $\theta^{C}_{33}$ can amplify the instability of the citrate formulation. It would be better to set $\theta^{X_f}_{33}$ and $\theta^{C}_{33}$ as the baseline value, i.e., zero. Similar to the general nonlinear pKG model, there are many less important policy parameters in the linear Gaussian pKG model (e.g., $\theta^{C}_{17}$ and $\theta^{N}_{33}$), which can be considered in dimension reduction.

\section{Conclusion}
\label{sec: conclusion}
In this study, we
propose an efficient SV-based risk and sensitivity analysis approach on pKG models, which can provide process risk and scientific understanding, and guide decision-making and stability control of biomanufacturing processes. We first present the method of SV estimation for the random factors, policy parameters, and model parameters in the general nonlinear pKG models, which can reduce the computation cost by reusing the pathway calculation in the pKG models. 
When the biomanufacturing process is monitored on a faster time scale, a linear Gaussian pKG model can be used to approximate it. Thus, we introduce the SV analysis for the linear Gaussian pKG model, in which the computation efficiency can be improved by exploring the linear Gaussian structure. 
We also provide a permutation sampling method, which is computationally efficient and can improve the estimation accuracy of the SV-based sensitivity analysis. Finally, we show how our method can help with interpretable sensitivity analysis in biomanufacturing using real experimental data and provide insights for the decision-making.
Future study can consider the potential effects of the interaction of the inputs in biomanufacturing. Notice that the SV estimation is computationally expensive, the approaches of the sensitivity analysis in parallel computing environments are also worth future investigation.

%
%
%





\bibliographystyle{informs2014}
\bibliography{sn-bibliography}

\clearpage
\renewcommand{\thepage}{ec\arabic{page}}  
\setcounter{page}{1}

\begin{APPENDICES}

	

	\vspace*{0.1cm}
	\begin{center}
		\large\textbf{Online Appendix --- ``Sensitivity Analysis on Policy-Augmented Graphical Hybrid Models with Shapley Value Estimation''}\\
	\end{center}
	\vspace*{0.3cm}

	
	\section{Technical Proofs}\label{secA: proof}
	\subsection{Proof of Proposition~\ref{prop: properties}}\label{subsec: proof_properties}
	Following Proposition 3.1 from \cite{castro2009polynomial}, for any given model parameters $\pmb{w}$,
	\begin{equation*}   \hat{\mbox{Sh}}\left(Y\vert {o},\pmb{w}\right) =\dfrac{1}{D}\sum_{d=1}^{D}\left[ \hat{g}(P_o(\pi^{(d)})\cup \{o\}\vert \pmb{w}) - \hat{g}(P_o(\pi^{(d)})\vert \pmb{w}) \right],
	\end{equation*}
	is an unbiased and consistent estimator of $\mbox{Sh}\left(Y\vert {o},\pmb{w}\right)$, which is defined in Equation~(\ref{eqn: sv_perm}). As $\mbox{Sh}\left(Y\vert {o}\right) = \mbox{E}_{\pmb{w}}\left[\mbox{Sh}\left(Y\vert {o},\pmb{w}\right)\right]$, we can simply see
	\begin{equation*}   \hat{\mbox{Sh}}\left(Y\vert {o}\right) =\dfrac{1}{QD}\sum_{q=1}^{Q}\sum_{d=1}^{D}\left[ \hat{g}(P_o(\pi^{(d)})\cup \{o\}\vert \pmb{w}^{(q)}) - \hat{g}(P_o(\pi^{(d)})\vert \pmb{w}^{(q)}) \right]
	\end{equation*}
	is an unbiased and consistent estimator of $\mbox{Sh}\left(Y\vert {o}\right)$. For each sample $\hat{g}(P_o(\pi^{(d)})\cup \{o\}\vert \pmb{w}^{(q)}) - \hat{g}(P_o(\pi^{(d)})\vert \pmb{w}^{(q)})$, the sample variance is 
	\begin{align*}
		\sigma^2 &= \mbox{E}_{\pmb{w}^{(q)},\pi^{(d)}}\left\{\left[\hat{g}(P_o(\pi^{(d)})\cup \{o\}\vert \pmb{w}^{(q)}) - \hat{g}(P_o(\pi^{(d)})\vert \pmb{w}^{(q)})-\mbox{Sh}\left(Y\vert {o}\right)\right]^2\right\} \\
		&= \frac{1}{\vert \mathcal{O}\vert!}\sum_{\pi\in \Pi(\mathcal{O})}\mbox{E}_{\pmb{w}}\left\{\left[\hat{g}(P_o(\pi)\cup \{o\}\vert \pmb{w}) - \hat{g}(P_o(\pi)\vert \pmb{w})-\mbox{Sh}\left(Y\vert {o}\right)\right]^2\right\}.
	\end{align*}
	As $\hat{\mbox{Sh}}\left(Y\vert {o}\right)$ is the average of $QD$ samples, the variance of $\hat{\mbox{Sh}}\left(Y\vert {o}\right)$ is $\dfrac{\sigma^2}{QD}$. For each sampled permutation $\pi^{(d)}$, we denote the set of variables that precede the $i$th element as $\pi_i^{(d)}$.  Therefore, 
	\begin{align*}
		\sum_{o\in\mathcal{O}}\hat{\mbox{Sh}}\left(Y\vert {o}\right) &=\dfrac{1}{QD}\sum_{q=1}^{Q}\sum_{d=1}^{D}\sum_{o\in \mathcal{O}}\left[ \hat{g}(P_o(\pi^{(d)})\cup \{o\}\vert \pmb{w}^{(q)}) - \hat{g}(P_o(\pi^{(d)})\vert \pmb{w}^{(q)}) \right]\\
		& = \dfrac{1}{QD}\sum_{q=1}^{Q}\sum_{d=1}^{D}\sum_{i=1}^{\vert\mathcal{O}\vert}\left[ \hat{g}(P_i(\pi^{(d)})\cup \{o\}\vert \pmb{w}^{(q)}) - \hat{g}(P_i(\pi^{(d)})\vert \pmb{w}^{(q)}) \right]\\
		&=\dfrac{1}{QD}\sum_{q=1}^{Q}\sum_{d=1}^{D}\left[ \hat{g}(\mathcal{O}\vert \pmb{w}^{(q)}) - \hat{g}(\emptyset\vert \pmb{w}^{(q)}) \right].
	\end{align*}
	
	\vspace{14pt}
	\subsection{Proof of Proposition~\ref{prop: sample_size}}\label{subsecA: proof_sample_size}
	By Chebyshev’s inequality, we have
	\begin{equation}
		\begin{aligned}
			\Pr(\vert\bar{X}-\mathbb{E}(X)\vert\ge r\sqrt{
				\mbox{Var}(X)}) \le \dfrac{1}{r^2},\nonumber
		\end{aligned}
	\end{equation}
	where $r$ is a constant. Let $\hat{\phi}_i$ be the $i$th marginal contribution sample of the input $o$ and $Z = \sum^{QD}_{i=1} \hat{\phi}_i$. Since $\hat{Sh}\left(Y\vert {o}\right) = Z/(QD)$, we have
	\begin{equation}
		\begin{aligned}
			\Pr(\vert Z-\mathbb{E}(Z)\vert\ge r\sqrt{\mbox{Var}(Z)}) = \Pr\left(\bigg\vert\dfrac{Z}{QD}-Sh\left(Y\vert {o}\right)\bigg\vert\ge \dfrac{r}{QD}\sqrt{\mbox{Var}(Z)}\right)\le\dfrac{1}{r^2}.\nonumber
		\end{aligned}
	\end{equation}
	Let $\epsilon=\dfrac{r}{QD}\sqrt{\mbox{Var}(Z)}$, we have
	\begin{equation}
		\begin{aligned}
			\Pr(\vert\hat{Sh}\left(Y\vert {o}\right)-Sh\left(Y\vert {o}\right)\vert\ge \epsilon) \le \dfrac{\mbox{Var}(Z)}{Q^2D^2\epsilon^2} = \dfrac{\mbox{Var}(\hat\phi_1)+\mbox{Var}(\hat\phi_2)+\cdots+\mbox{Var}(\hat\phi_{QD})}{Q^2D^2\epsilon^2} = \dfrac{\sigma^2}{QD\epsilon^2}.\nonumber
		\end{aligned}
	\end{equation}
	If our aim is to bound the left-hand side of the above inequality to be at most $\delta$,
	we can set $QD \ge \lceil \dfrac{\sigma^2}{\delta\epsilon^2} \rceil$.
	
	If $Z$ is the sum of $QD$
	independent random variables, i.e., $Z = X_1 + X_2 + \cdots +X_{QD}$, and $X_i$ is bounded by two values, $a_i$ and $b_i$, i.e., $a_i \le X_i \le b_i$. Then by Hoeffding's inequality, 
	the following result holds:
	\begin{equation}
		\begin{aligned}
			\Pr(\vert Z-\mathbb{E}(Z)\vert\ge t) \le 2\exp(-\dfrac{2t^2}{\sum_{i=1}^{QD}(b_i-a_i)^2}).\nonumber
		\end{aligned}
	\end{equation}
	Therefore, if we define $Z = \sum_{i=1}^{QD}\hat{\phi}_i$ and $r = b_i - a_i$, where $\hat{\phi}_i$ is the $i$th marginal contribution sample of the input $o$, we have
	\begin{equation}
		\begin{aligned}
			\Pr(\vert Z-QD(Sh\left(Y\vert {o}\right))\vert \ge t) =  \Pr(\vert\hat{Sh}\left(Y\vert {o}\right)-Sh\left(Y\vert {o}\right)\vert\ge \dfrac{t}{QD})\le 2\exp(-\dfrac{2t^2}{QDr^2}).\nonumber
		\end{aligned}
	\end{equation}
	Let $\epsilon = \dfrac{t}{QD}$, we have
	\begin{equation}
		\begin{aligned}
			\Pr(\vert\hat{Sh}\left(Y\vert {o}\right)-Sh\left(Y\vert {o}\right)\vert\ge \epsilon) \le 2\exp(-\dfrac{2QD\epsilon^2}{r^2}).\nonumber
		\end{aligned}
	\end{equation}
	If our aim is to bound the left-hand side of the above inequality to be at most $\delta$,
	it follows that $QD \ge \lceil \dfrac{\ln(2/\delta)r^2}{2\epsilon^2} \rceil$.
	\vspace{14pt}

	\subsection{Proof of Theorem \ref{thm: predictive SV for random factors}}\label{subsecA: proof_them1}
	Let $\mathcal{O}$ be the set containing all random factors, the SV of $o = e_h^k$ given model parameters $\pmb{w}$ can be derived as
	\begin{equation}
		\begin{aligned}
			\mbox{Sh}\left(\pmb{s}_{t+1}\vert o,\pmb{\theta}; \pmb{w}\right)
			&=\sum_{\mathcal{U}\subset \mathcal{O}/\{{o}\}}\dfrac{(\vert\mathcal{O}\vert -\vert\mathcal{U}\vert-1)! \vert\mathcal{U}\vert!} {\vert\mathcal{O}\vert!}\left[ g(\mathcal{U}\cup\left\{{o}\right\}) - g(\mathcal{U}) \right]\\
			&=\sum_{\mathcal{U}\subset \mathcal{O}/\{{o}\}}\dfrac{(\vert\mathcal{O}\vert -\vert\mathcal{U}\vert-1)! \vert\mathcal{U}\vert!} {\vert\mathcal{O}\vert!} \left[\mathbb{E}(\pmb{s}_{t+1}\vert\mathcal{U}\cup\{o\},\pmb{\theta};\pmb{w})\right.\\
			&\quad\quad\left.-\mathbb{E}(\pmb{s}_{t+1}\vert\mathcal{U}, \pmb{\theta};\pmb{w})\right]
			\\
			&=\mathbf{R}_{h,t}e_h^k\mathrm{1}_k, \nonumber
		\end{aligned}
	\end{equation}
	\begin{equation}
		\begin{aligned}
			\mbox{Sh}\left(r_{t+1}\vert{o},\pmb{\theta}; \pmb{w}\right)
			&=\sum_{\mathcal{U}\subset \mathcal{O}/\{{o}\}}\dfrac{(\vert\mathcal{O}\vert -\vert\mathcal{U}\vert-1)! \vert\mathcal{U}\vert!} {\vert\mathcal{O}\vert!}\left[ g(\mathcal{U}\cup\left\{{o}\right\}) - g(\mathcal{U}) \right]\\
			&=\sum_{\mathcal{U}\subset \mathcal{O}/\{{o}\}}\dfrac{(\vert\mathcal{O}\vert -\vert\mathcal{U}\vert-1)! \vert\mathcal{U}\vert!} {\vert\mathcal{O}\vert!} \left[\mathbb{E}(r_{t+1}\vert\mathcal{U}\cup\{o\},\pmb{\theta};\pmb{w})\right.\\
			&\quad\quad\left.-\mathbb{E}(r_{t+1}\vert\mathcal{U}, \pmb{\theta};\pmb{w})\right]
			\\
			&=(\pmb{b}_{t+1}^{\top}\pmb{\theta}_{t+1}^\top+\pmb{c}_{t+1}^\top)\mathbf{R}_{h,t}e_h^k\mathrm{1}_k. \nonumber
		\end{aligned}
	\end{equation}
	
	\vspace{14pt}
	
	\subsection{Proof of Theorem \ref{thm: variance-based SV for random factors}}\label{subsecA: proof_them2}
	Let $\mathrm{1}_{\mathcal{U}_{\pmb{X}_{t+1}}}$ denote the vector which has the same length with $\pmb{X}_{t+1}$.
	If the $k$th element of $\pmb{X}_{t+1}$ is in $\mathcal{U}$, the $k$th element of $\mathrm{1}_{\mathcal{U}_{\pmb{X}_{t+1}}}$ is set to be 1. Otherwise, it is set to be 0.
	For the variance of the state $\pmb{s}_{t+1}$,
	\begin{align*}
		g(\mathcal{U}\vert\pmb{w})&=\mathbb{E}\left[\mbox{Var}\left[\pmb{s}_{t+1}\vert\mathcal{O}/\mathcal{U},\pmb{\theta}; \pmb{w}\right]\right]\\
		&=\mathbf{R}_{t+1}\left[\mathbf{V}_{t+1}\odot(\mathrm{1}_{\mathcal{U}_{\pmb{X}_{t+1}}}\mathrm{1}_{\mathcal{U}_{\pmb{X}_{t+1}}}^{\top})\right]\mathbf{R}_{t+1}^{\top},
	\end{align*}
	where $\odot$ denotes component-wise multiplication. Therefore, if $o$ is the $l$th entry of $\pmb{X}_{t+1}$,
	\begin{align}
		&\mbox{Sh}\left( \pmb{s}_{t+1}\vert {o},\pmb{\theta};\pmb{w}\right)
		=\sum_{\mathcal{U}\subset \mathcal{O}/\{{o}\}}\dfrac{(\vert\mathcal{O}\vert-\vert\mathcal{U}\vert-1)! \vert\mathcal{U}\vert!} {\vert\mathcal{O}\vert!}\left[ g(\mathcal{U}\cup\left\{{o}\right\}) - g(\mathcal{U}) \right]\nonumber\\
		&= \sum_{k=0}^{\vert\mathcal{O}\vert-1}\sum_{\substack{\vert\mathcal{U}\vert=k \nonumber\\ \mathcal{U}\subset \mathcal{O}/\{{o}\}}}\dfrac{(\vert\mathcal{O}\vert -\vert\mathcal{U}\vert-1)! \vert\mathcal{U}\vert!} {\vert\mathcal{O}\vert!}\left[ g(\mathcal{U}\cup\left\{{o}\right\}) - g(\mathcal{U}) \right]\nonumber\\
		&= \sum_{k=0}^{\vert\mathcal{O}\vert-1}\sum_{\substack{\vert\mathcal{U}\vert=k \nonumber\\ \mathcal{U}\subset \mathcal{O}/\{{o}\}}}\dfrac{(\vert\mathcal{O}\vert -\vert\mathcal{U}\vert-1)! \vert\mathcal{U}\vert!} {\vert\mathcal{O}\vert!}\mathbf{R}_{t+1}\left[\mathbf{V}_{t+1}\odot \Delta_o(\mathrm{1}_{\mathcal{U}_{\pmb{X}_{t+1}}}\mathrm{1}_{\mathcal{U}_{\pmb{X}_{t+1}}}^{\top})\right]\mathbf{R}_{t+1}^{\top}\nonumber\\
		&= \sum_{k=0}^{\vert\mathcal{O}\vert-1}\mathbf{R}_{t+1}\left[\mathbf{V}_{t+1}\odot \left[\sum_{\substack{\vert\mathcal{U}\vert=k \nonumber\\ 
				\mathcal{U}\subset \mathcal{O}/\{{o}\}}}\dfrac{(\vert\mathcal{O}\vert -\vert\mathcal{U}\vert-1)! \vert\mathcal{U}\vert!} {\vert\mathcal{O}\vert!} {\Delta}_{o}(\mathrm{1}_{\mathcal{U}_{\pmb{X}_{t+1}}}\mathrm{1}_{\mathcal{U}_{\pmb{X}_{t+1}}}^{\top})\right]\right]\mathbf{R}_{t+1}^{\top}\nonumber\\
		&= \mathbf{R}_{t+1}\left[\mathbf{V}_{t+1} \odot \left[\frac{1}{2}(\mathbf{1}_{t+1}^{l}+\mathbf{1}_{t+1}^{l\top})\right]\right]\mathbf{R}_{t+1}^{\top}\nonumber,
	\end{align}
	where $\mathbf{1}_{t+1}^{l}$ is the matrix whose shape is the same as $\mathbf{V}_{t+1}$ and the entries in $l$th column are 1 and other entries in other columns are 0, ${\Delta}_{o}(\mathrm{1}_{\mathcal{U}_{\pmb{X}_{t+1}}}\mathrm{1}_{\mathcal{U}_{\pmb{X}_{t+1}}}^{\top})$ is the matrix whose  $l$th row and $l$th column is $\mathrm{1}_{\mathcal{U}_{\pmb{X}_{t+1}}}$ except the $(l, l)$th entry is 1, other elements not included in $l$th row or $l$th column are 0. 
	
	For the cumulative reward, let $\pmb{\alpha}_{t+1}=\pmb{b}_{t+1}^\top\pmb{\theta}_{t+1}^\top+\pmb{c}_{t+1}^\top$, we have
	\begin{align*}
		&\mbox{Var}(r_{t+1}\vert\mathcal{O}/\mathcal{U},\pmb{\theta};\pmb{w})=\mbox{Var}\Biggl\{\pmb{\alpha}_{t+1}\left[\sum_{i=1}^{t}\mathbf{R}_{i,t}\pmb{e}_i + \pmb{e}_{t+1}\right]\Big\vert\mathcal{O}/\mathcal{U},\pmb{\theta};\pmb{w}\Biggr\}\\
		&=\mbox{Var}(\pmb{\alpha}_{t+1}\mathbf{R}_{t+1}\pmb{X}_{t+1}\odot\mathrm{1}_{\mathcal{U}_{\pmb{X}_{t+1}}})\\
		&=\pmb{\alpha}_{t+1}\mathbf{R}_{t+1}\left[\mathbf{V}_{t+1}\odot(\mathrm{1}_{\mathcal{U}_{\pmb{X}_{t+1}}}\mathrm{1}_{\mathcal{U}_{\pmb{X}_{t+1}}}^{\top})\right]\mathbf{R}_{t+1}^{\top}\pmb{\alpha}_{t+1}^{\top},
	\end{align*}
	\begin{align*}
		&\mbox{Cov}(r_{t_1+1}, r_{t_2+1}\vert\mathcal{O}/\mathcal{U},\pmb{\theta};\pmb{w})=\mbox{Cov}\Biggl\{\pmb{\alpha}_{t_1+1}\Biggl[\sum_{i=1}^{t_1}\mathbf{R}_{i,t_1}\pmb{e}_i+ \pmb{e}_{t_1+1}\Biggr],\\
		&\quad\pmb{\alpha}_{t_2+1}\left[\sum_{i=1}^{t_2}\mathbf{R}_{i,t_2}\pmb{e}_i + \pmb{e}_{t_2+1}\right]\Big\vert\mathcal{O}/\mathcal{U},\pmb{\theta};\pmb{w}\Biggr\}\\
		&=\mbox{Cov}(\pmb{\alpha}_{t_1+1}\mathbf{R}_{t_1+1}\pmb{X}_{t_1+1}\odot\mathrm{1}_{\mathcal{U}_{\pmb{X}_{t_1+1}}},   \pmb{\alpha}_{t_2+1}\mathbf{R}_{t_2+1}\pmb{X}_{t_2+1}\odot\mathrm{1}_{\mathcal{U}_{\pmb{X}_{t_2+1}}})\\
		&=\pmb{\alpha}_{t_1+1}\mathbf{R}_{t_1+1}\left[\mathbf{V}_{t_1+1, t_2+1}\odot(\mathrm{1}_{\mathcal{U}_{\pmb{X}_{t_1+1}}}\mathrm{1}_{\mathcal{U}_{\pmb{X}_{t_2+1}}}^{\top})\right]\mathbf{R}_{t_2+1}^{\top}\pmb{\alpha}_{t_2+1}^{\top},
	\end{align*}
	where $t_1 < t_2$ and $\mathbf{V}_{t_1+1, t_2+1}$ is the covariance matrix between $\pmb{X}_{t_1+1}$ and $\pmb{X}_{t_2+1}$. Therefore, for the cumulative reward,
	\begin{align*}
		&g(\mathcal{U}\vert\pmb{w})=\mathbb{E}\left[\mbox{Var}\left[\sum_{t=0}^{H-1} r_{t+1}\mid\mathcal{O}/\mathcal{U},\pmb{\theta};\pmb{w}\right]\right] \\
		&=\sum_{t=0}^{H-1}\pmb{\alpha}_{t+1}\mathbf{R}_{t+1}\left[\mathbf{V}_{t+1}\odot(\mathrm{1}_{\mathcal{U}_{\pmb{X}_{t+1}}}\mathrm{1}_{\mathcal{U}_{\pmb{X}_{t+1}}}^{\top})\right]\mathbf{R}_{t+1}^{\top}\pmb{\alpha}_{t+1}^{\top}\nonumber\\
		&+2\sum_{t_2=1}^{H-1}\sum_{t_1=0}^{t_2-1}\pmb{\alpha}_{t_1+1}\mathbf{R}_{t_1+1}\left[\mathbf{V}_{t_1+1, t_2+1}\odot(\mathrm{1}_{\mathcal{U}_{\pmb{X}_{t_1+1}}}\mathrm{1}_{\mathcal{U}_{\pmb{X}_{t_2+1}}}^{\top})\right]\mathbf{R}_{t_2+1}^{\top}\pmb{\alpha}_{t_2+1}^{\top}.
	\end{align*}
	If $o$ is the $l$th entry of $\pmb{X}_{H}$ and is in period $h$, then 
	\begin{align*}
		&\mbox{Sh}\left(\sum_{t=0}^{H-1} r_{t+1}\vert{o},\pmb{\theta};\pmb{w}\right)
		=\sum_{\mathcal{U}\subset \mathcal{O}/\{{o}\}}\dfrac{(\vert\mathcal{O}\vert -\vert\mathcal{U}\vert-1)! \vert\mathcal{U}\vert!} {\vert\mathcal{O}\vert!}\left[ g(\mathcal{U}\cup\left\{{o}\right\}) - g(\mathcal{U}) \right]\nonumber\\
		&=\sum_{k=0}^{\vert\mathcal{O}\vert-1}\sum_{\substack{\vert\mathcal{U}\vert=k \\ \mathcal{U}\subset \mathcal{O}/\{{o}\}}}\dfrac{(\vert\mathcal{O}\vert -\vert\mathcal{U}\vert-1)! \vert\mathcal{U}\vert!} {\vert\mathcal{O}\vert!}\left[ g(\mathcal{U}\cup\left\{{o}\right\}) - g(\mathcal{U}) \right]\nonumber\\
		&=\sum_{k=0}^{\vert\mathcal{O}\vert-1}\sum_{\substack{\vert\mathcal{U}\vert=k \nonumber\\ \mathcal{U}\subset \mathcal{O}/\{{o}\}}}\dfrac{(\vert\mathcal{O}\vert -\vert\mathcal{U}\vert-1)! \vert\mathcal{U}\vert!} {\vert\mathcal{O}\vert!}\left[\sum_{t=h-1}^{H-1}\pmb{\alpha}_{t+1}\mathbf{R}_{t+1}\left[\mathbf{V}_{t+1}\odot \Delta_o(\mathrm{1}_{\mathcal{U}_{\pmb{X}_{t+1}}}\mathrm{1}_{\mathcal{U}_{\pmb{X}_{t+1}}}^{\top})\right]\right.\nonumber\\
		&\left.\mathbf{R}_{t+1}^{\top}\pmb{\alpha}_{t+1}^{\top}+2\sum_{t_2=h-1}^{H-1}\sum_{t_1=0}^{t_2-1}\pmb{\alpha}_{t_1+1}\mathbf{R}_{t_1+1}\left[\mathbf{V}_{t_1+1,t_2+1}\odot\Delta_o(\mathrm{1}_{\mathcal{U}_{\pmb{X}_{t_1+1}}}\mathrm{1}_{\mathcal{U}_{\pmb{X}_{t_2+1}}}^{\top})\right]\mathbf{R}_{t_2+1}^{\top}\pmb{\alpha}_{t_2+1}^{\top}\right]\nonumber\\
		&=\sum_{t=h-1}^{H-1}\pmb{\alpha}_{t+1}\mathbf{R}_{t+1}\left[\mathbf{V}_{t+1} \odot \left[\frac{1}{2}(\mathbf{1}_{t+1}^{l}+\mathbf{1}_{t+1}^{l\top})\right]\right]\mathbf{R}_{t+1}^{\top}\pmb{\alpha}_{t+1}^{\top}\\
		&\quad+\sum_{t_2=h-1}^{H-1}\sum_{t_1=0}^{h-2}\pmb{\alpha}_{t_1+1}\mathbf{R}_{t_1+1}\left[\mathbf{V}_{t_1+1, t_2+1}\odot(\dfrac{1}{2}\mathbf{1}_{t_1+1, t_2+1}^{l})\right]\mathbf{R}_{t_2+1}^{\top}\pmb{\alpha}_{t_2+1}^{\top}\\
		&\quad+\sum_{t_2=2}^{H-1}\sum_{t_1=h-1}^{t_2-1}\pmb{\alpha}_{t_1+1}\mathbf{R}_{t_1+1}\left[\mathbf{V}_{t_1+1, t_2+1}\odot\left[\dfrac{1}{2}(\mathbf{1}_{t_1+1, t_2+1}^{l}+\mathbf{1}_{t_2+1, t_1+1}^{l\top})\right]\right]\mathbf{R}_{t_2+1}^{\top}\pmb{\alpha}_{t_2+1}^{\top},
	\end{align*}
	where $\mathbf{1}_{t_1+1, t_2+1}^{l}$ is the matrix whose shape is the same as $\mathbf{V}_{t_1+1, t_2+1}$ and the entries in the $l$th column are 1 and other entries in other columns are 0. 
	
	\vspace{14pt}
	
	\subsection{Proof of Proposition \ref{prop: time complexity analysis for predictive SV}}\label{subsecA: proof_prop5}
	
	Note that the model parameters $\pmb\beta_{t}^s \in \mathbb{R}^{n\times n}$, $\pmb\beta_{t}^a \in \mathbb{R}^{m\times n}$, $\pmb\theta_{t} \in \mathbb{R}^{n\times m}$, $\pmb{b}_{t} \in \mathbb{R}^{m}$, $\pmb{c}_{t} \in \mathbb{R}^{n}$, $\pmb{\mu}_{t}^s \in \mathbb{R}^{n}$.
	Recall that the naive multiplication of two matrices $A \in \mathbb{R}^{n \times m}$ and $B \in \mathbb{R}^{m \times p}$ requires $O(nmp)$ time. Therefore, the update of the $\mathbf{R}_{1,t}
	=\left[\left(\pmb\beta_{t}^s\right)^\top + \left(\pmb\beta_{t}^a\right)^\top\pmb\theta_{t}^\top\right]\mathbf{R}_{1,t-1}$ requires $O(n^3+n^2m)$,
	the calculation of $\mathbf{R}_{1,t}(\pmb{s}_{0} - \pmb\mu^{s}_{1})$ takes $O(n^2)$ time.  Therefore, the calculation of the value function takes $O(H(n^3+n^2m+n^2)) = O(H(n^3+n^2m))$. As the number of the policy parameters is $Hnm$, the calculation of the SV estimation requires $O(QDHnm(H(n^3+n^2m))) = O(QDH^2(n^4m+n^3m^2))$. For the brute force algorithm, we calculate $\mathbf{R}_{1,t}$ using the formula directly, which costs $O(t(n^3+n^2m))$ and the overall calculation of the SV requires $O(QDH^3(n^4m+n^3m^2))$. Therefore, we can reduce the computational complexity by a factor of $O(H)$ compared to a brute force approach that does not reuse in the pathway calculation.
	
	\vspace{14pt}
	
	\subsection{Proof of Proposition \ref{prop: time complexity analysis for variance-based SV}}\label{subsecA: proof_prop6}
	Note that the model parameters $\pmb\beta_{t}^s \in \mathbb{R}^{n\times n}$, $\pmb\beta_{t}^a \in \mathbb{R}^{m\times n}$, $\pmb\theta_{t} \in \mathbb{R}^{n\times m}$, $\pmb{b}_{t} \in \mathbb{R}^{m}$, $\pmb{c}_{t} \in \mathbb{R}^{n}$. Therefore, the calculation of $\mathbf{R}_{t,t}=(\pmb\beta_{t}^s+\pmb\theta_{t}\pmb\beta_{t}^a)^\top$ takes $O(n^2m+n^3)$ time. Therefore the computational time in step 1 of Algorithm~\ref{algovarsv1} is $ O(H^2(n^2m+n^3))$.  
	The calculation of  $\pmb{\alpha}_{t_1+1}\mbox{Cov}(\pmb{s}_{t_1+1},\pmb{s}_{t_2+1})\pmb{\alpha}_{t_2+1}^{\top}$ requires $O(n^2)$ time, so the calculation in the step 2 of Algorithm~\ref{algovarsv1} requires $O(H^2n^2)$. 
	As a result, the total computational cost of the Algorithm~\ref{algovarsv1} is $O(H^2(n^2m+n^3))$.
	The overall calculation of the variance-based SV requires $O(QD\vert \mathcal{O}\vert H^2(n^2m+n^3))=O(QDH^3(n^4m+n^3m^2))$. For the brute force algorithm, which calculates the covariance matrix directly, the calculation of $\mbox{Cov}(\pmb{s}_{t_1+1},\pmb{s}_{t_2+1})$ requires $O(H^3(n^2m+n^3))$. The total computational cost becomes $O(H^6(n^4m+n^3m^2))$. Therefore, we can reduce the computational complexity by a factor of $O(H^3)$ compared to a brute force approach.
	
	\vspace{14pt}

	\subsection{Proof of Proposition \ref{prop: time complexity analysis for permutation sampling}}\label{subsecA: proof_prop4}
	
	In BMT, the calculation of $\pmb{z}$ and $\pmb{x}$ cost $O(s)$. Therefore, the total cost to obtain $D$ points is $O(Ds)$. In TFWW transformation, the recursive procedure for the calculation of $\pmb{g}$ , $\pmb{d}$ and $\pmb{x}$ requires $O(s)$. Thus the total cost to obtain $D$ points is $O(Ds)$. In SCT method, the calculation of the vector $\pmb{\phi}$ using inverse c.d.f. takes $O(s)$ due to the
	calculation of c.d.f. in the root founding algorithm, which is shown as follows:
	\begin{align*}
		I_{n}&=\int \sin^{n}xdx \\
		&=\int \sin x \sin^{n-1}x dx\\
		&=-\cos x \sin ^{n-1} x+\int(n-1) \cos ^{2} x \sin^{n-1}x dx\\
		&=-\cos x \sin ^{n-1} x+(n-1) I_{n-2}-(n-1) I_{n}.
	\end{align*}
	Therefore we can write $I_{n}$ as
	\begin{equation}
		\begin{aligned}
			I_{n}=-\frac{1}{n} \cos x \sin ^{n-1} x+\frac{n-1}{n} I_{n-2}.
			\nonumber
		\end{aligned}
	\end{equation}
	As $I_{n} = x$ when $n=0$ and $I_{n} = 1-\cos x$ when $n=1$, we can compute the c.d.f. recursively and the time complexity is $O(s)$. As such calculation occurs for each dimension of each sample, the total computational complexity of spherical coordinate transformation is $O(Ds^2)$. 

	\section{Hypersphere Sampling Procedure}\label{secA1}
	Here we describe the BMT and SCT procedures, as well as TFWW transformation,  to sample uniformly distributed points on the unit hypersphere. For more details, interested readers may refer to \cite{mitchell2022sampling} and \cite{fang1993number}.

	\subsection{Box-Muller Transformation (BMT) in \cite{mitchell2022sampling}}
	\label{agm: BMT}
	For the BMT method, let $a$ be the integer satisfying  $s \leq 2 a \leq s+1$, \\
	\textbf{Step 1}. Generate $n$ uniformly distributed points from $C^{(2a)}$:\\
	$\left\{\boldsymbol{c}^{(k)}=\left(c^{(k)}_{1},
	\cdots, c^{(k)}_{2a}\right), k=1, \cdots, D\right\}$;\\
	\textbf{Step 2}. For $k=1,\cdots, D$, set
	\begin{equation}
		\begin{aligned}
			Z^{(k)}_{2 i-1} & =\sqrt{-2 \log c^{(k)}_{2 i-1}} \cos \left(2 \pi c^{(k)}_{2 i}\right) ,\nonumber\\
			Z^{(k)}_{2 i} & =\sqrt{-2 \log c^{(k)}_{2 i-1}} \sin \left(2 \pi c^{(k)}_{2 i}\right), \quad i=1, \cdots, a,\nonumber
		\end{aligned}
	\end{equation}
	where   \cite{muller1959note} prove that $\pmb{z}^{(k)}=\left(Z^{(k)}_{1}, \cdots, Z^{(k)}_{s}\right)^{\top} \sim N_{s}\left(\mathbf{0}, \pmb{I}_{s}\right)$, the $s$-dimensional standard normal distribution; \\
	\textbf{Step 3}. For $k=1,\cdots,D$, calculate
	$\pmb{x}^{(k)}=\frac{\pmb{z}^{(k)}}{\sqrt{\pmb{z}^{(k)\prime}\pmb{z}^{(k)}}}\nonumber$.

Then $\left\{x^{(k)}=\left(x^{(k)}_{1}, \cdots, x^{(k)}_{s}\right), k=1, \cdots, D\right\}$ is uniformly distributed over  $U_{s}$ \citep{muller1959note}.

\vspace{14pt}

\subsection{Spherical Coordinate Transformation (SCT) in \cite{mitchell2022sampling}}
\label{agm: SCT}
For the SCT method, \\
\textbf{Step 1}. Generate $n$ uniformly distributed points on $C^{(s-1)}$:\\
$\left\{\boldsymbol{c}^{(k)}=\left(c^{(k)}_{1},
\cdots, c^{(k)}_{s-1}\right), k=1, \cdots, D\right\}$;\\
\textbf{Step 2}. For $k=1, \cdots, D$, $i = 1, \cdots, s-1$:\\
Solve the root of $c^{(k)}_{i} - \frac{\pi}{B\left(\frac{1}{2}, \frac{s-j}{2}\right)} \int_{0}^{\phi}(\sin (\pi t))^{s-j-1} dt$ using bracketing method and obtain $\phi^{(k)}_{i}$;\\
\textbf{Step 3}. For $k=1, \cdots, D$, calculate\\
\begin{equation}
\begin{aligned}
	x^{(k)}_{s} =\prod_{i=1}^{s-1} S^{(k)}_{i}; \quad x^{(k)}_{j} =\prod_{i=1}^{j-1} S^{(k)}_{i} A^{(k)}_{j}, \quad j=1, \cdots, s-1, \nonumber
\end{aligned}
\end{equation}
where
\begin{equation}
\begin{aligned}
	\begin{array}{l}
		S^{(k)}_{i}=\sin \left(\pi \phi^{(k)}_{i}\right), \quad A^{(k)}_{i}=\cos \left(\pi \phi^{(k)}_{i}\right), \quad i=1, \cdots, s-2, \nonumber\\
		S^{(k)}_{s-1}=\sin \left(2 \pi \phi^{(k)}_{s-1}\right), \quad A^{(k)}_{s-1}=\cos \left(2 \pi \phi^{(k)}_{s-1}\right).\nonumber
	\end{array}
\end{aligned}
\end{equation}

Then $\left\{x^{(k)}=\left(x^{(k)}_{1}, \cdots, x^{(k)}_{s}\right), k=1, \cdots, D\right\}$ is uniformly distributed over  $U_{s}$. 

\vspace{14pt}

\subsection{TFWW Transformation in \cite{fang1993number}}
\label{subsec: Appendix_tfww}
\subsubsection{The dimension of the hypersphere is even.}
When the dimension of the hypersphere $s=2a$  is even and  $t_{1}=t_{2}=\cdots=t_{a}=  2$, we have  $\pmb{x}_j \sim \mathrm{Unif}\left(U_{2}\right)$  and  $\left(d_{1}^{2}, \cdots, d_{a}^{2}\right) \sim D_{a}(1, \cdots, 1)=   \mathrm{Unif}\left(T_{a}\right)$, where $T_{s}=\left\{\left(x_{1}, \cdots, x_{s}\right) \in R_{+}^{s}: x_{1}+\cdots+x_{s}=1\right\}$. Now
\begin{equation}
\begin{aligned}
	\pmb{x} \stackrel{d}{=}\left(\begin{array}{c}
		d_{1} \cos \left(2 \pi \phi_{1}\right) \\
		d_{1} \sin \left(2 \pi \phi_{1}\right) \\
		d_{2} \cos \left(2 \pi \phi_{2}\right) \\
		d_{2} \sin \left(2 \pi \phi_{2}\right) \\
		\vdots \\
		d_{a} \cos \left(2 \pi \phi_{a}\right) \\
		d_{a} \sin \left(2 \pi \phi_{a}\right)
	\end{array}\right),   
	\nonumber
\end{aligned}
\end{equation}
where
\begin{itemize}    
\item[(1)]  $\left(d_{1}^{2}, \cdots, d_{a}^{2}\right) \sim \mathrm{Unif}\left(T_{a}\right)$  and  $d_{j}>0, j=1, \cdots, a$.
\item[(2)]  $\left(\phi_{1}, \cdots, \phi_{a}\right) \sim \mathrm{Unif}\left(C^{a}\right)$.
\item[(3)]  $\left(\phi_{1}, \cdots, \phi_{a}\right)$  and  $\left(d_{1}, \cdots, d_{a}\right)$  are independent.

\end{itemize}    
With this fact, we have the following algorithm:\\
\textbf{Step 1}. Generate $\left\{c^{(k)}=\left(c^{(k)}_{1}, \cdots, c^{(k)}_{s-1}\right), k=1, \cdots, D\right\}$ on $C^{(s-1)}$ uniformly.\\
\textbf{Step 2}. Set  $g^{(k)}_{a}=1$  and  $g^{(k)}_{0}=0, k=1, \cdots, D$.\\
\textbf{Step 3}. Recursively compute for  $k=1, \cdots, D$,
\begin{equation}
\begin{aligned}
	g^{(k)}_{j}=g^{(k)}_{j+1} (c^{(k)}_{j})^{1 / j}\text{, } j=a-1, a-2, \cdots, 1.
	\nonumber
\end{aligned}
\end{equation}
\textbf{Step 4}. Compute $d^{(k)}_{l}=\sqrt{g^{(k)}_{l}-g^{(k)}_{l-1}}$ and 
\begin{equation}
\begin{aligned}
	\left\{\begin{array}{ll}
		x^{(k)}_{2 l-1}=d^{(k)}_{l} \cos \left(2 \pi, c^{(k)}_{a+l-1}\right), & l=1, \cdots, a; \\
		x^{(k)}_{2 l}=d^{(k)}_{l} \sin \left(2 \pi, c^{(k)}_{a+l-1}\right), & k=1, \cdots, D.
	\end{array}\right.
	\nonumber
\end{aligned}
\end{equation}

Then  $\left\{\pmb{x}^{(k)}=\left(x^{(k)}_{ 1}, \cdots, x^{(k)}_{s}\right), k=1, \cdots, D\right\}$ is uniformly distributed on $U_{s}$.

\subsubsection{The dimension of the hypersphere is odd.}
When the dimension of the hypersphere $s=2t+3 $ $(t \geq 0)$  is odd, we choose  $t_{1}=3, t_{2}=\cdots=   t_{a}=2$, and  $a=t+1$, and we have  $\left(d_{1}^{2}, \cdots, d_{a}^{2}\right) \sim   D_{a}(3 / 2,1, \cdots, 1)$. Therefore, we obtain
\begin{equation}
\begin{aligned}
	\left(d_{1}^{2}, \cdots, d_{a}^{2}\right) \stackrel{d}{=}\left(\prod_{i=1}^{t} B_{i},\left(1-B_{1}\right) \prod_{i=2}^{t} B_{i}, \cdots, 1-B_{t}\right),
	\nonumber
\end{aligned}
\end{equation}
where  $B_{1}, \cdots, B_{t}$  are independent,  $B_{i}$ follows beta distribution with parameters $(2 i+1) / 2$ and $1$, $i=   1, \cdots, t$. It is easy to see that the c.d.f. of  $B_{i}$  is  $F(x) = x^{(2 i+1) / 2}, \text{ } 0<x<1$. The stochastic representation now can be expressed as
\begin{equation}
\begin{aligned}
	\boldsymbol{x} \stackrel{d}{=}\left(\begin{array}{c}
		d_{1}\left(1-2 \phi_{1}\right) \\
		d_{1} \sqrt{\phi_{1}\left(1-\phi_{1}\right)} \cos \left(2 \pi \phi_{2}\right) \\
		d_{1} \sqrt{\phi_{1}\left(1-\phi_{1}\right)} \sin \left(2 \pi \phi_{2}\right) \\
		d_{2} \cos \left(2 \pi \phi_{3}\right) \\
		d_{2} \sin \left(2 \pi \phi_{3}\right) \\
		\vdots \\
		d_{a} \cos \left(2 \pi \phi_{a+1}\right) \\
		d_{a} \sin \left(2 \pi \phi_{a+1}\right)
	\end{array}\right),
	\nonumber
\end{aligned}
\end{equation}

where
\begin{itemize}  
\item[(1)]  $\left(d_{1}^{2}, \cdots, d_{a}^{2}\right) \sim D_{a}(3 / 2,1, \cdots, 1)$.
\item[(2)]  $\left(\phi_{1}, \cdots, \phi_{a+1}\right) \sim \mathrm{Unif}\left(C^{a+1}\right)$.
\item[(3)]  $\left(\phi_{1}, \cdots, \phi_{a+1}\right)$, and  $\left(d_{1}^{2}, \cdots, d_{a}^{2}\right)$  are independent.

\end{itemize}  

Therefore we have the following algorithm:\\
\textbf{Step 1}. Generate $n$ uniformly distributed points on $C^{(s-1)}$:\\
$\left\{\boldsymbol{c}^{(k)}=\left(c^{(k)}_{1},
\cdots, c^{(k)}_{s-1}\right), k=1, \cdots, D\right\}$;\\
\textbf{Step 2}. Set  $a=(s-1) / 2$, $g^{(k)}_{a}=1$  and  $g^{(k)}_{ 0}=0, k=1, \cdots, D$;\\
\textbf{Step 3}. Recursively compute for  $k=1, \cdots, D$ 
\begin{equation}
\begin{aligned}
	g^{(k)}_{j}=g^{(k)}_{j+1} (c^{(k)}_{ j})^{2 /(2 j+1)}, \quad j=a-1, a-2, \cdots, 1;
	\nonumber
\end{aligned}
\end{equation}
\textbf{Step 4}. Compute $d^{(k)}_{ j}=\sqrt{g^{(k)}_{j}-g^{(k)}_{j-1}}$, $\quad j=1, \cdots, a, k=1, \cdots, D$.\\
\textbf{Step 5}. For  $k=1, \cdots, D$,  compute
\begin{equation}
\begin{aligned}
	x^{(k)}_{1} & =d^{(k)}_{1}\left(1-2 c^{(k)}_{a}\right), \\
	x^{(k)}_{2} & =d^{(k)}_{1} \sqrt{c^{(k)}_{a}\left(1-c^{(k)}_{a}\right)} \cos \left(2 \pi c^{(k)}_{a+1}\right), \\
	x^{(k)}_{3} & =d^{(k)}_{1} \sqrt{c^{(k)}_{a}\left(1-c^{(k)}_{a}\right)} \sin \left(2 \pi c^{(k)}_{a+1}\right), \\
	x^{(k)}_{2 l} & =d^{(k)}_{l} \cos \left(2 \pi c^{(k)}_{2 l}\right), \\
	x^{(k)}_{2 l+1} & =d^{(k)}_{l} \sin \left(2 \pi c^{(k)}_{2 l}\right), \quad l=2,3, \cdots, a.
	\nonumber
\end{aligned}
\end{equation}

Then  $\left\{\pmb{x}^{(k)}=\left(x^{(k)}_{1}, \cdots, x^{(k)}_{s}\right), k=1, \cdots, D\right\}$ are uniformly distributed on  $U_{s}$.



\vspace{20pt}



\section{SV Estimation Procedures for Policy Parameters in Linear Gaussian pKG}\label{sec: agm_policy}

In this section, we present the detailed SV analysis procedures for policy parameters in the linear Gaussian pKG model. Algorithm~\ref{agm: sv_linear_predictive} illustrates the predictive SV analysis, where the calculation reusing method from Equation~(\ref{eq: nested_propogation}) is reflected in lines~16-21.
Algorithm~\ref{agm: sv_linear_variance} illustrates the variance-based SV analysis, which utilizes Algorithm~\ref{algovarsv1}
in line 16 to compute the variance with calculation reusing.

\SetInd{0.5em}{1em}
\SetNlSkip{1em}
\begin{algorithm}[h!]
\linespread{1.2}\selectfont
Input: the number of samples for model parameters $Q$, the number of permutations $D$, the initial state value $\pmb{s}_0$, the policy parameters $\pmb{\theta}$, the distribution of model parameters $p(\pmb{w}\vert \mathcal{D})$, the set of inputs $\mathcal{O} =\{\theta_t^{ij},1\le i\le n, 1\le j \le m, 1\le t\le H\}$,
$\hat{Sh}^{r}_o  = 0 \text{ for } o \in \mathcal{O}$ and $1\le t\le H$.
\\
Generate the samples of the marginal contribution and calculate the sum of samples:\\
\For{$q = 1,2,\ldots,Q$}{
	Sample model parameters from the posterior distribution:\\ 
	$\pmb{w}^{(q)}\sim p(\pmb{w}\vert\mathcal{D})$;\\
	\For{$d=1,2,\ldots,D$}{
		Sample the random permutation $\pi^{(d)}$ from $\Pi(\mathcal{O})$ with Algorithm~\ref{algosampling};\\
		\For{$i=0,1,\ldots,\vert \mathcal{O}\vert$}{
			$\bar{\pmb\theta} = \pmb \theta$;\\
			\eIf{$i = 0$}{$\mathcal{U} = \emptyset$, $\bar{\pmb\theta}_{\mathcal{O}/\mathcal{U}}=0$, $\hat{g}_{prev}^{r} = 0$, $1\le t\le H$;}{$\mathcal{U} = \mathcal{U}\cup\{\pi^{(d)}(i)\}$, where $\pi^{(d)}(i)$ is the $i$th element of $\pi^{(d)}$;
				\\ $\bar{\pmb\theta}_{\mathcal{O}/\mathcal{U}}=0$, $\hat{g}^{r}_{prev}=\hat{g}^{r}$, $1\le t\le H$;}
			$\pmb{R}_{1,0} = \pmb{\mbox{I}}$;\\
			\For{$t=1,\ldots, H-1$}{
				$\mathbf{R}_{1,t}=\left[\left(\pmb\beta_{t}^{s(q)}\right)^\top + \left(\pmb\beta_{t}^{a(q)}\right)^\top\bar{\pmb\theta}_{t}^\top\right]\mathbf{R}_{1,t-1}$
			}
			$\hat{g}^{r} = \sum_{t=0}^{H-1}\pmb\alpha_{t+1}\left[\mathbf{R}_{1,t}(\pmb{s}_{0} - \pmb\mu^{s}_{1})\right]$, $\hat{Sh}^{r}_{\pi^{(d)}(i)} = \hat{Sh}^{r}_{\pi^{(d)}(i)}+\hat{g}^{r}-\hat{g}^{r}_{prev}$;}}}
Calculate the estimation of the SV:\\
$\hat{Sh}^{r}_o = \hat{Sh}^{r}_o/(QD) \text{ for } o \in \mathcal{O}$.
\caption{Predictive SV for Policy Parameters of Linear Gaussian pKG}\label{agm: sv_linear_predictive}
\end{algorithm}

\SetInd{0.5em}{1em}
\SetNlSkip{1em}
\begin{algorithm}[!t]
\linespread{1.2}\selectfont
Input: the number of samples for model parameters $Q$, the number of permutations $D$, the initial state value $\pmb{s}_0$, the policy parameters $\pmb{\theta}$, the distribution of model parameters $p(\pmb{w}\vert \mathcal{D})$, the set of inputs $\mathcal{O} =\{\theta_t^{ij},1\le i\le n, 1\le j \le m, 1\le t\le H\}$,
$\hat{Sh}^{r}_o  = 0 \text{ for } o \in \mathcal{O}$ and $1\le t\le H$.
\\
Generate the samples of the marginal contribution and calculate the sum of samples:\\
\For{$q = 1,2,\ldots,Q$}{
	Sample model parameters from the posterior distribution:\\ 
	$\pmb{w}^{(q)}\sim p(\pmb{w}\vert\mathcal{D})$;\\
	\For{$d=1,2,\ldots,D$}{
		Sample the random permutation $\pi^{(d)}$ from $\Pi(\mathcal{O})$ with Algorithm~\ref{algosampling};\\
		\For{$i=0,1,\ldots,\vert \mathcal{O}\vert$}{
			$\bar{\pmb\theta} = \pmb \theta$;\\
			\eIf{$i = 0$}{$\mathcal{U} = \emptyset$, $\bar{\pmb\theta}_{\mathcal{O}/\mathcal{U}}=0$, $\hat{g}_{prev}^{r} = 0$, $1\le t\le H$;}{$\mathcal{U} = \mathcal{U}\cup\{\pi^{(d)}(i)\}$, where $\pi^{(d)}(i)$ is the $i$th element of $\pi^{(d)}$;
				\\ $\bar{\pmb\theta}_{\mathcal{O}/\mathcal{U}}=0$, $\hat{g}^{r}_{prev}=\hat{g}^{r}$, $1\le t\le H$;}
			$\hat{g}^{r} = \mbox{Var}(r)$, calculated by Algorithm~\ref{algovarsv1};\\
			$\hat{Sh}^{r}_{\pi^{(d)}(i)} = \hat{Sh}^{r}_{\pi^{(d)}(i)}+\hat{g}^{r}-\hat{g}^{r}_{prev}$;}}}
Calculate the estimation of the SV:\\
$\hat{Sh}^{r}_o = \hat{Sh}^{r}_o/(QD) \text{ for } o \in \mathcal{O}$.
\caption{Variance-based SV for Policy Parameters of Linear Gaussian pKG}\label{agm: sv_linear_variance}
\end{algorithm}

\section{Kinetic Equations}\label{secA2}
\begin{table}[ht]
\linespread{1}\selectfont
\centering
\caption{Model parameters of kinetic equations.}\label{tab:model parameters}
\vspace{0.3cm}
\begin{tabular}{lp{8.5cm}ll}
	\hline
	\multicolumn{1}{l}{Parameters} & \multicolumn{1}{l}{Descrption} & \multicolumn{1}{l}{Estimation} &\multicolumn{1}{l}{Unit} \\
	\hline
	$\alpha_L$ & Coefficient of lipid production for cell growth & 0.1273 & - \\
	$C_{max}$ & Maximum citrate concentration that cells can tolerate & 130.90 &  g/L \\
	$K_{iN}$ & Nitrogen limitation constant to trigger on lipid and citrate production & 0.1229 &  g/L \\
	$K_{iS}$ &Inhibition constant for substrate in lipid-based growth kinetics &  612.18 &  g/L \\
	$K_{iX}$ & Constant for cell density effect on cell growth and lipid/citrate formation &  59.974 & g/L \\
	$K_N$ & Saturation constant for intracellular nitrogen in growth kinetics & 0.0200 & g/L \\
	$K_O$ & Saturation constant for dissolved oxygen in kinetics of cell growth,
	substrate uptake, lipid consumption by $\beta$-oxidation & 0.3309 & \% \\
	$K_S$ & Saturation constant for substrate utilization &  0.0430 &  g/L \\
	$K_{SL}$ & Coefficient for lipid consumption/decomposition & 0.0217 & - \\
	$m_s$ & Maintenance coefficient for substrate & ------ &  g/g/h \\
	$r_L$ & Constant ratio of lipid carbon flow to total carbon flow (lipid + citrate) & ------ & - \\
	$V_{evap}$ & Evaporation rate (or loss of volume) in the fermentation & 0.0026 &  L/h \\
	$Y_{cs}$ & Yield coefficient of citrate based on substrate consumed &  0.6826 & g/g \\
	$Y_{ls}$ & Yield coefficient of lipid based on substrate consumed & 0.3574 & g/g \\
	$Y_{xn}$ & Yield coefficient of cell mass based on nitrogen consumed &  10.0 & g/g \\
	$Y_{xs}$ & Yield coefficient of cell mass based on substrate consumed &  0.2386 & g/g \\
	$\beta_{LC_{max}}$& Coefficient of maximum carbon flow for citrate and lipid &  ------ &  h-1 \\
	$\mu_{max}$ & Maximum specific growth rate on substrate &  ------ & h-1 \\
	$S_F$ & Oil concentration in oil feed & 917.00 & g/L \\
	\hline
\end{tabular}
\end{table}%

\begin{enumerate}

\item  Cell mass: The total cell mass $X$ consists of lipid-free ($X_f$) and lipid ($L$) mass, and is measured by
dry cell weight (DCW) in fermentation experiments:
\begin{equation*}
	X = X_f + L.
\end{equation*}

\item  Dilution rate: The dilution $D$ of the working liquid volume in the bioreactor is caused by feed of base,
such as KOH solution ($F_B$) and substrate ($F_S$):
\begin{equation*}
	D = \frac{F_B + F_S}{V}.
\end{equation*}

\item Lipid-free cell growth: Cell growth consumes nutrients, including the substrate (carbon source) $S$,
nitrogen $N$, and dissolved oxygen $O$, and is described by coupled Monod equations, with considerations of
inhibitions from high oil concentrations and cell densities:
\begin{align*}
	dX_f &= \mu\cdot X_f dt - \left(D - \frac{V_{evap}}{V}\right)X_f dt + \sigma(X_f)dB_t,\\
	\mu &= \mu_{max}\left(\frac{S}{K_S + S}\cdot\frac{1}{1 + S/K_{iS}}\right)\cdot\frac{N}{K_N + N}\cdot\frac{O}{K_O + O}\cdot\frac{1}{1 + X_f/K_{ix}}.
\end{align*}

\item Citrate accumulation: Citrate ($C$) is an overflow of all the carbon introduced to the lipid synthesis
pathway ($\beta_{LC}$), which is more active under nitrogen-limited, substrate-rich, and aerobic conditions. Only a
proportion $r_L$ of the total carbon flow (citrate plus lipid) in the lipid synthesis pathway goes to production due to the overflow loss in citrate. A tolerance limit $C_{max}$ of citrate, and the effect of cell density on product
formation $(1/(1 + X_f /K_{iX}))$, are also considered in the model.
\begin{align*}
	dC &= \beta_C \cdot X_f dt - \left(D - \frac{V_{evap}}{V}\right)C dt + \sigma(C)dB_t,\\
	\beta_C &= 2(1-r_L)\beta_{LC},\\
	\beta_{LC} &= \frac{1}{1+N/K_{iN}}\cdot \left(\frac{S}{K_S+S}\cdot\frac{1}{1+S/K_{iS}}\right)\frac{O}{K_O+O}\cdot \frac{1}{1+X_f/K_{iX}}
	\nonumber\\
	& \quad\left(1-\frac{C}{C_{max}}\right)\beta_{LC_{max}}.
\end{align*}

\item Lipid accumulation: Lipid is accumulated under nitrogen-limited, substrate-rich, and aerobic conditions. Lipid production is described using partial growth-association kinetics; a small portion of lipid can be
degraded when its concentration is high in the presence of oxygen.
\begin{align*}
	dL &= q_L \cdot X_f dt - \left(D - \frac{V_{evap}}{V}\right)L dt + \sigma(L)dB_t,\\
	q_L&= \alpha_L\cdot\mu+r_L\cdot\beta_{LC}-K_{SL}\frac{L}{L+X_f}\cdot\frac{O}{K_O+O}.
\end{align*}

\item Substrate consumption: Oil ($S$) is fed during the fed-batch fermentation and used for cell growth,
energy maintenance, citrate formation, and lipid production.
\begin{align*}
	-dS &= q_S \cdot X_f dt -\frac{F_S}{V}S_Fdt + \left(D - \frac{V_{evap}}{V}\right)S dt + \sigma(S)dB_t,\\
	q_S&= \frac{1}{Y_{X/S}}\mu + \frac{O}{K_O+O}\cdot \frac{S}{K_S+S}m_S+\frac{1}{Y_{C/S}}\beta_C+\frac{1}{Y_{L/S}}\beta_{L}.
\end{align*}

\item Nitrogen consumption: Extracellular nitrogen ($N$) is used for cell growth.
\begin{equation*}
	-dN = \frac{1}{Y_{X/N}}\mu X_f dt + \left(D - \frac{V_{evap}}{V}\right)Ndt + \sigma(N)dB_t.
\end{equation*}

\item Volume change: The rate $V$ of working volume change of the fermentation is calculated based on the
rates of base feed ($F_B$), substrate feed ($F_S$), and evaporation ($V_{evap}$).
\begin{align*}
	dV &= (F_B + F_S - V_{evap}) dt + \sigma(V)dB_t,\\
	F_B&= \frac{V}{1000}\left(\frac{7.14}{Y_{X/N}}\mu X_f + 1.59\beta_C X_f\right).
\end{align*}

\end{enumerate}

\end{APPENDICES}

\end{document}